\documentclass[11pt]{article}

\usepackage[final]{acl}
\usepackage{float}
\usepackage{placeins}
\usepackage{times}
\usepackage{latexsym}

\usepackage[T1]{fontenc}
\usepackage{tabularx}

\usepackage{tablefootnote}

\usepackage[utf8]{inputenc}

\usepackage{microtype}

\usepackage{inconsolata}

\usepackage{graphicx}

\title{Less is MoE: Trimming Experts in Domain-Specialist Language Models}

\author{Haoze He$^{1}$\thanks{Equal Contribution} \enspace
Xinkai Zou$^{2}$\footnotemark[1] \enspace
Xuan Jiang$^{3}$ \enspace
Xingyuan Ding$^{1}$ \enspace \\
\textbf{Ao Qu}$^{3}$ \enspace
\textbf{Juncheng Billy Li}$^{1}$ \quad
\textbf{Heather Miller}$^{1}$ \\[6pt]
$^{1}$Carnegie Mellon University \quad
$^{2}$UCSD \quad
$^{3}$MIT \quad \\
\texttt{\{haozeh, xingyuad, junchenl, heather.miller\}@cs.cmu.edu} \\
\texttt{jayzou@ucsd.edu} \quad
\texttt{\{xuanj, qua\}@mit.edu} \\
\textbf{Project Page:}~\url{https://less-is-moe.github.io/} \\
\textbf{Code Repo:}~\url{https://github.com/HectorHHZ/Less-is-MoE}
}

\usepackage{url}
\usepackage{amsfonts}
\usepackage{amssymb}
\usepackage{amsmath}
\usepackage{mathtools}
\usepackage{amsthm}
\usepackage{nicefrac}
\usepackage{dsfont}
\newcommand{\mathbbm}[1]{\mathds{#1}}
\usepackage[table]{xcolor}
\usepackage{subcaption}
\usepackage{booktabs}
\usepackage{caption}
\usepackage{adjustbox}
\usepackage{soul}
\usepackage{multirow}
\usepackage[ruled,vlined,linesnumbered]{algorithm2e}
\usepackage[capitalize,noabbrev]{cleveref}
\usepackage{enumitem}
\usepackage[most]{tcolorbox}

\newtcolorbox{examplecard}{
  colback=gray!8,         
  colframe=gray!25,       
  boxrule=0.3pt,
  arc=2pt,                
  left=8pt, right=8pt, top=6pt, bottom=6pt,
  before skip=6pt, after skip=6pt,
}

\newcommand{\hK}{\cellcolor{blue!10}}   
\newcommand{\hC}{\cellcolor{blue!14}}   
\newcommand{\hM}{\cellcolor{blue!18}}   
\newcommand{\hR}{\cellcolor{blue!22}}   
\newcommand{\hA}{\cellcolor{red!28}}    

\newcolumntype{K}{>{\columncolor{blue!2}\centering\arraybackslash}X}
\newcolumntype{O}{>{\columncolor{blue!8}\centering\arraybackslash}X}
\newcolumntype{T}{>{\columncolor{blue!14}\centering\arraybackslash}X}
\newcolumntype{R}{>{\columncolor{blue!20}\centering\arraybackslash}X}
\newcolumntype{V}{>{\columncolor{red!12}\centering\arraybackslash}X}
\newcolumntype{D}{>{\centering\arraybackslash}X}     

\begin{document}

\maketitle

\begin{abstract}

Mixture-of-Experts (MoE) models achieve strong performance through conditional computation, but their large parameter footprint poses deployment challenges. Prior MoE compression approaches catastrophically fail when evaluated on general-purpose benchmarks beyond commonsense reasoning. We trace this failure to the granularity of compression: important capabilities are distributed across experts but concentrated in FFN sparse intermediate dimensions. To identify these dimensions, we use Fisher importance which outperforms activation-, router-score-, and magnitude-based alternatives, and identifies tiny sets of task-critical dimensions: in Qwen1.5-MoE, removing as few as 12 of 1.35M routed-FFN intermediate dimensions collapses \texttt{GSM8K} accuracy while largely preserving factual-knowledge performance.
Building on this, we propose \textbf{Fisher-MoE}, which operates within FFN to remove intermediate dimensions ranked by Fisher importance. At the same 50\% MoE compression ratio, Fisher-MoE preserves model capability, while reducing weight memory by $~\sim$45\% and improving inference throughput by 21\%.
These findings suggest intermediate dimension granularity is an effective unit for both compression and ranking where capability concentrates in MoE models.

\end{abstract}

\section{Introduction}
\label{sec:introduction}

Mixture-of-Experts (MoE) models have emerged as a dominant paradigm for scaling language model capacity while maintaining efficient inference through conditional computation~\citep{shazeer2017outrageously, lepikhin2021gshard, fedus2022switch}. This enables models with tens of billions of parameters to achieve efficient inference, but their large total parameter footprint still poses significant challenges for deployment in terms of memory, storage, and serving.

To reduce this footprint, prior work compresses MoE models by removing or merging experts based on heuristic importance metrics, such as activation frequency~\citep{muzio2024seer, lu-etal-2024-experts, chen2022task}, router scores~\citep{xie2024moe, gu2025delta}, or weight magnitudes~\citep{lee2024stun, yang-etal-2024-moe, li2023merge, chen2024retraining}. Despite differences in these metrics, existing approaches share a common design choice: compression is performed at the granularity of entire experts. Moreover, existing methods are primarily evaluated on commonsense reasoning benchmarks, which we find to be unstable and weak indicators of compression quality.(see Appendix~\ref{appendix:random_prune_variance}) We instead evaluate on more challenging general-purpose benchmarks spanning mathematical reasoning, code generation, knowledge, and multilingual understanding following the evaluation settings in official technical reports~\citep{qwen_moe, yang2025qwen3}. The picture changes dramatically. Under a fair controlled comparison using a unified compression~\citet{he2025towards} framework, we evaluate activation-, score-, and magnitude-based methods at a fixed MoE compression ratio $p=50\%$ (the fraction of routed-expert FFN parameters removed; defined in \S\ref{subsec:compression_formulation}). As shown in Figure~\ref{fig:fisher_vs_others}, all existing expert-level approaches suffer catastrophic performance collapse on benchmarks such as \texttt{GSM8K}, \texttt{HumanEval}, \texttt{MBPP}, and \texttt{MATH}.

\begin{figure}[t]
\centering
\includegraphics[width=\columnwidth]{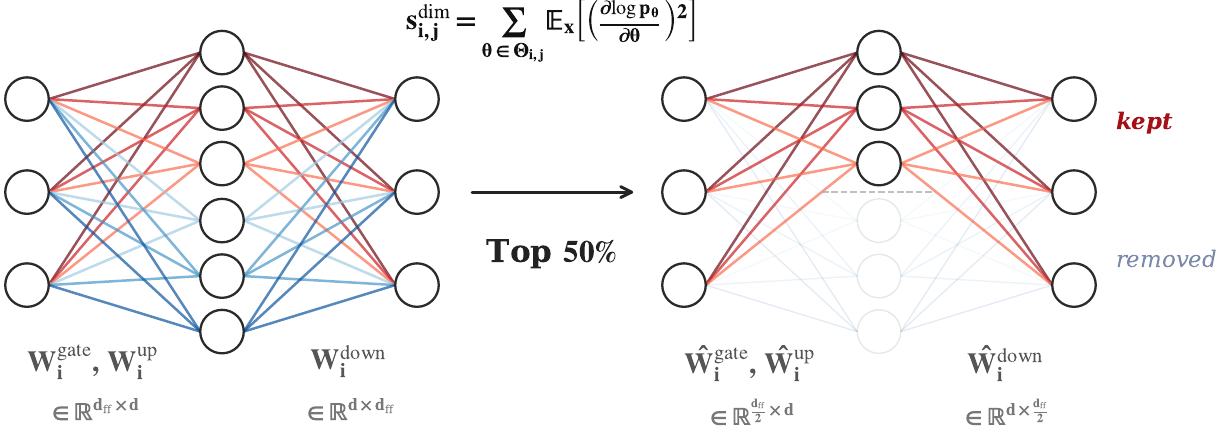}

\caption{Intermediate dimension compression of a single MoE expert FFN.                 
Edge colors encode the Fisher importance score $s_{i,j}^{\mathrm{dim}}$ of   
each intermediate dimension (red = high, blue = low).                              
The bottom $50\%$ of dimensions by Fisher score are removed (\emph{faded}),
reducing $W_i^{\mathrm{gate}}, W_i^{\mathrm{up}} \in \mathbb{R}^{d_{\mathrm{ff}} \times d}$                                                                                     
and $W_i^{\mathrm{down}} \in \mathbb{R}^{d \times d_{\mathrm{ff}}}$                     
to $\hat{d}_{\mathrm{ff}} = d_{\mathrm{ff}}/2$ without discarding any expert.}

\label{fig:intermediate_dim_compression}
\end{figure}

We trace the failure of existing MoE compression methods to two factors. First, inaccurate importance metrics fail to estimate the importance of the parameters. Second, compression at an overly coarse granularity assumes capability is localized at the expert level, whereas in reality it is distributed across experts but concentrated in a small subset of intermediate dimensions. 
Our contribution is not the Fisher information itself but rather the choice of attributable intermediate dimension unit in an MoE, and aggregates parameter-level Fisher scores into intermediate dimension scores of experts. We use Fisher importance as the underlying scoring tool to localize this attribution (\S\ref{sec:gradient_metrics}) and to design a finer-grained compression method at the intermediate dimension level (\S\ref{sec:intermediate_compression}).

\paragraph{(1) Fisher importance as a unit attribution tool.}
Prior methods rely on activation ratios, weight magnitude, or router scores, which we show are poor proxies for parameter importance. In contrast, empirical Fisher information performs substantially better as both a compression metric and an attribution tool. We validate this through three lines of evidence: (a) Fisher importance outperforms existing metrics under controlled comparison, (b) zeroing just 12 out of 1.35 million intermediate dimensions identified by Fisher importance destroys mathematical reasoning while preserving general knowledge, and (c) removing the bottom 50\% of dimensions preserves overall performance (\S\ref{sec:gradient_metrics}).


\paragraph{(2) Fine-grained intermediate dimension compression.}
Prior methods operate at the level of entire experts. However, capability in MoE models is distributed across experts but concentrated in a small subset of intermediate dimensions. Removing entire experts therefore discards critical dimensions alongside redundant ones. We close this expert-vs-intermediate-dimension localization gap by proposing \textbf{Fisher-MoE}, which converts the per-dimension attribution above into a structurally smaller MoE: it performs fine-grained compression within each expert by physically resizing the rows of $W^{\text{gate}}, W^{\text{up}}$ and the columns of $W^{\text{down}}$ corresponding to low-Fisher-score dimensions (\S\ref{sec:intermediate_compression}).

Our key contributions are:

\begin{itemize}[leftmargin=*]
    \item We report a structural property of MoE models in which capability is not localized at the expert level but instead concentrated in a small subset of intermediate dimensions distributed across experts. 
    This suggests one reason for the failure of expert-level MoE compression methods.
    

    \item We define the intermediate dimension as the attributable structure 
    and use Fisher information as a tool for characterizing this structure. We empirically show that Fisher importance identifies critical and redundant intermediate dimension. 
    \item We propose \textbf{Fisher-MoE}, a fine-grained compression method that operates at the intermediate-dimension level instead of expert-level. 
    At a $p=50\%$ compression ratio, it preserves downstream performance while improving inference throughput by 21\%.

\end{itemize}

\section{Model Capability Attribution}
\label{sec:gradient_metrics}
Can we find an importance metric that ranks parameters by their impact on model capability? In this section, we suggest the \emph{empirical Fisher information} as a better metric. We first derive Fisher importance from this expansion and contrast it with prior heuristics (\S\ref{subsec:formalize_metrics}) then provide three lines of empirical evidence that this score is a useful ranking signal: it outperforms all alternatives (\S\ref{para:gradient_comparison}), masking few Fisher ranked critical dimensions collapses generation-heavy tasks while sparing knowledge tasks (\S\ref{sec:critical_dimensions}), and the set of Fisher-important dimensions shared across tasks is small and removing it collapses every task (\S\ref{subsec:shared_dimensions}). For brevity, we group benchmarks into four categories: Knowledge covers \texttt{MMLU}, \texttt{CEval} and \texttt{CMMLU}; Code covers \texttt{HumanEval} and \texttt{MBPP}; Reasoning is \texttt{BBH}; Math covers \texttt{MATH} and \texttt{GSM8K}.

\subsection{Empirical Fisher Information}
\label{subsec:formalize_metrics}

Let $p_\theta(y \mid x)$ be the model's output distribution and $\mathcal{D}$ a calibration dataset of $N$ samples with sequence length $T$. We denote by $\mathcal{L}(x,y) = -\log p_\theta(y\mid x)$ the loss. Each expert $E_i$ is a gated FFN
\begin{equation}
E_i(\mathbf{x}) = W_i^{\text{down}} \left(\sigma(W_i^{\text{gate}} \mathbf{x}) \odot W_i^{\text{up}} \mathbf{x}\right),
\end{equation}
with $W_i^{\text{gate}}, W_i^{\text{up}} \in \mathbb{R}^{d_{\text{ff}} \times d}$, $W_i^{\text{down}} \in \mathbb{R}^{d \times d_{\text{ff}}}$, and $\sigma(\cdot)$ an element-wise nonlinearity (e.g., SiLU).

\paragraph{Empirical Fisher approximation.}
A natural way to ask ``how important is parameter $\theta_i$?'' is to ask how much the predictive distribution moves when we perturb it. For a small $\delta \in \mathbb{R}^{|\theta|}$, a second-order Taylor expansion of the KL divergence between the unperturbed and perturbed models yields

\begin{equation}
\mathbb{E}_x \mathrm{KL}\!\left[p_\theta \,\Vert\, p_{\theta+\delta}\right] = \delta^\top F_\theta\, \delta + O(\|\delta\|^3),
\label{eq:kl_taylor}
\end{equation}
where $F_\theta \in \mathbb{R}^{|\theta|\times|\theta|}$ is the \emph{Fisher Information Matrix}, Letting $g_\theta(x,y) := \nabla_\theta \log p_\theta(y\mid x)$ denote the per-token score function,

\begin{equation}
F_\theta = \mathbb{E}_{x,y}\!\left[g_\theta(x,y)\, g_\theta(x,y)^{\top}\right].
\label{eq:full_fisher}
\end{equation}
The full $F_\theta$ is intractable for billion-parameter models. We adopt the \emph{empirical Fisher}: we replace $\mathbb{E}_{y\sim p_\theta(y\mid x)}$ by the ground-truths in $\mathcal{D}$, yielding
\begin{equation}
\hat F_\theta \;=\; \frac{1}{N}\sum_{(x,y)\in\mathcal{D}} \left(\nabla_\theta \log p_\theta(y\mid x)\right)^2
\label{eq:empirical_fisher}
\end{equation}
Throughout the rest of the paper we refer to this score simply as the \emph{Fisher}. When we apply this metric to MoE compression we call the resulting method Fisher-MoE.


\paragraph{Baselines.} Prior MoE compression instead uses one of three metrics:
\begin{itemize}[leftmargin=*]
    \item \emph{Activation ratio:} $s_i^{\text{act}} = \frac{1}{NT}\sum_{(\mathbf{x},t)\in\mathcal{D}} \mathbbm{1}[i\in T(\mathbf{x}_t)]$, the fraction of tokens routed to expert $i$~\citep{muzio2024seer, lu-etal-2024-experts}.
    \item \emph{Router score:} $s_i^{\text{score}} = \frac{1}{NT}\sum g_i(\mathbf{x}_t)$, the average gating weight~\citep{xie2024moe}.
    \item \emph{Magnitude:} $s_i^{\text{mag}} = \|W_i^{\text{gate}}\|_F + \|W_i^{\text{up}}\|_F + \|W_i^{\text{down}}\|_F$, a data-independent weight norm~\citep{lee2024stun, yang-etal-2024-moe}.
\end{itemize}

\paragraph{Extension to intermediate dimensions.}
A key practical advantage of Fisher importance is that the same derivation applies whether $W$ is a router weight matrix, a full expert FFN, or a single intermediate dimension. We instantiate Eq.~\ref{eq:empirical_fisher} at three granularities used throughout the paper:
\emph{\textbf{(1) Router-level Fisher}}: $W$ is the row of the router gate corresponding to expert $i$; \emph{\textbf{(2) Expert-level Fisher}}: $W$ is the union of $W_i^{\text{gate}}, W_i^{\text{up}}, W_i^{\text{down}}$ for expert $i$; \emph{\textbf{(3) Intermediate dimension Fisher}}: $W$ is the rows of $W_i^{\text{gate}}, W_i^{\text{up}}$ and the column of $W_i^{\text{down}}$ corresponding to a single intermediate dimension $j$ of expert $i$. $G_i^{m} := \nabla_{W_i^{m}}\!\mathcal{L}$ for $m\in\{g,u,d\}$,
\begin{equation}
    s_{i,j}^{\text{Fisher}} = \frac{1}{d}\sum_k \big[ (G_i^{g})_{j,k}^2 + (G_i^{u})_{j,k}^2 + (G_i^{d})_{k,j}^2 \big]
    \label{eq:dim_importance}
\end{equation}
averaged over calibration samples; the sum runs over the hidden dimension $k\in[d]$ and the denominator $d$ normalizes by the total number of parameters tied to dimension $j$. The intermediate dimension form is what enables both fine-grained compression (removing low-scoring dimensions, \S\ref{sec:intermediate_compression}) and fine-grained attribution (identifying high-scoring critical dimensions, \S\ref{sec:critical_dimensions}).

\begin{figure}[t]
\centering
\includegraphics[width=\columnwidth]{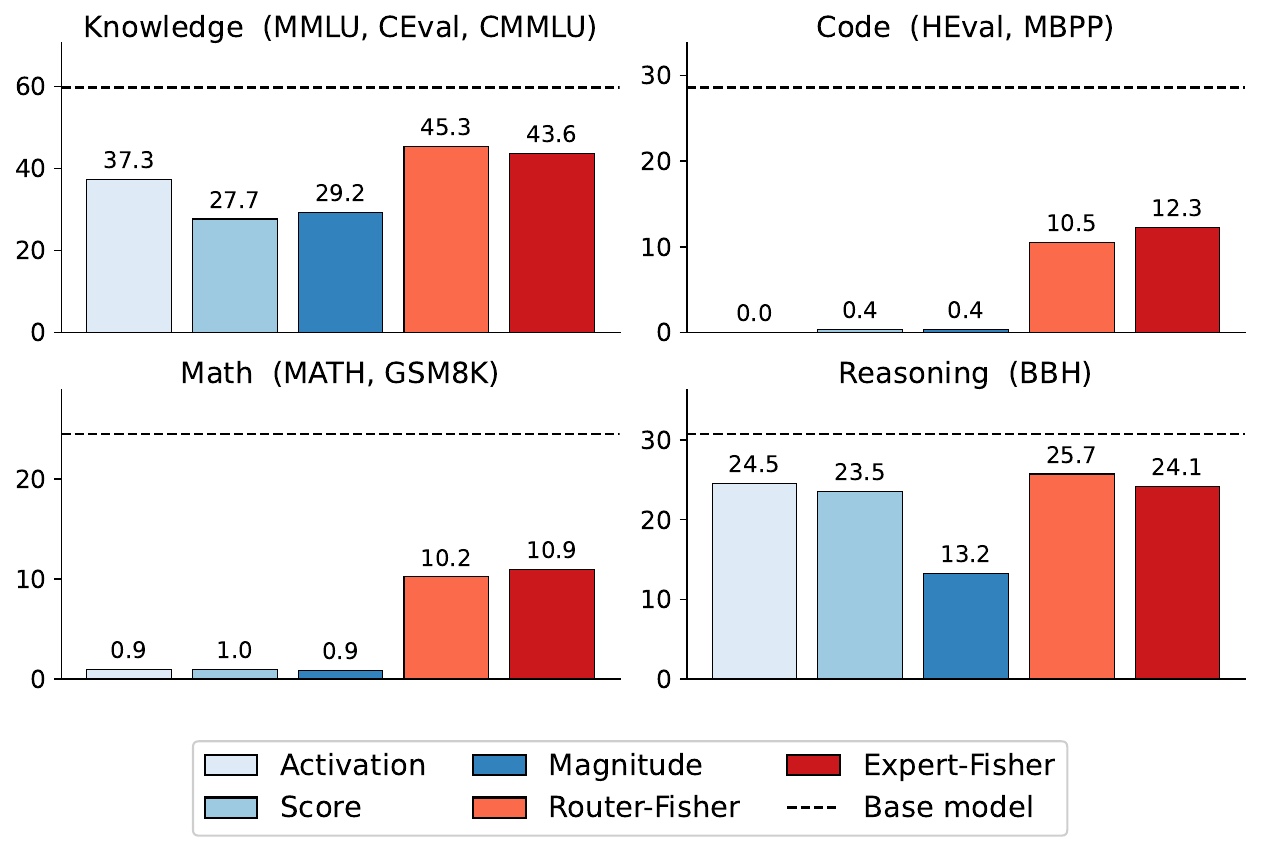}
\caption{Expert-level pruning with existing important metrics on Qwen1.5-MoE at $50\%$ compression ratio.}
\label{fig:fisher_vs_others}
\end{figure}

\paragraph{}
\label{para:gradient_comparison}
We compare all four importance metrics under expert-level compression at MoE compression ratio $p=50\%$. To ensure a fair comparison, we use the unified compression framework of~\citep{he2025towards}: all methods share identical calibration data (GSM8K training set, 128 samples), compression procedure, and evaluation protocol on Qwen1.5-MoE; only the importance metric differs. 
Fisher importance dominates across all tasks (Figure~\ref{fig:fisher_vs_others}).
Full results can be found in Table~\ref{tab:raw_fig2}.

\subsection{Can Fisher Importance Identify Task-Critical Parameters?}
\label{sec:critical_dimensions}
\begin{table}[h]
\centering
\small
\setlength{\tabcolsep}{4pt}
\resizebox{\columnwidth}{!}{%
\begin{tabular}{lccccc}
\toprule
\multirow{2}{*}{\textbf{Setting}} 
& \multirow{2}{*}{\textbf{Knowledge}} 
& \multirow{2}{*}{\textbf{Code}} 
& \multirow{2}{*}{\textbf{Reasoning}} 
& \multicolumn{2}{c}{\textbf{Math}} \\
\cmidrule(lr){5-6}
& & & & \textbf{MATH} & \textbf{GSM8K} \\
\midrule
\textbf{Base model}
& 59.8 & 28.6 & 30.7 & 13.0 & 35.9 \\
\midrule
\multicolumn{6}{l}{\textit{Top-masking}} \\
\textbf{Mask-Top-12}
& 56.7 & 10.7 & 23.6 
& \cellcolor{blue!8}1.1\
& \cellcolor{blue!8}0.8\ \\
\midrule
\multicolumn{6}{l}{\textit{Critical vs Redundant Dimension Removal}} \\
\textbf{Remove redundant} $(\mathcal{D}_{\cap})$
& 61.1 & 28.5 & 31.2 
& 8.7\
& 31.2\ \\
\textbf{Remove critical} $(\mathcal{K}_{\cap})$
& 1.7 & 0.0 & 0.2 
& 0.0\
& 0.0\ \\
\bottomrule
\end{tabular}%
}
\caption{Impact of masking top-12 and removing critical/redundant intermediate dimensions on performance.}
\label{tab:reverse_cross_domain}
\end{table}

\begin{figure}[t]
\centering
\includegraphics[width=\columnwidth]{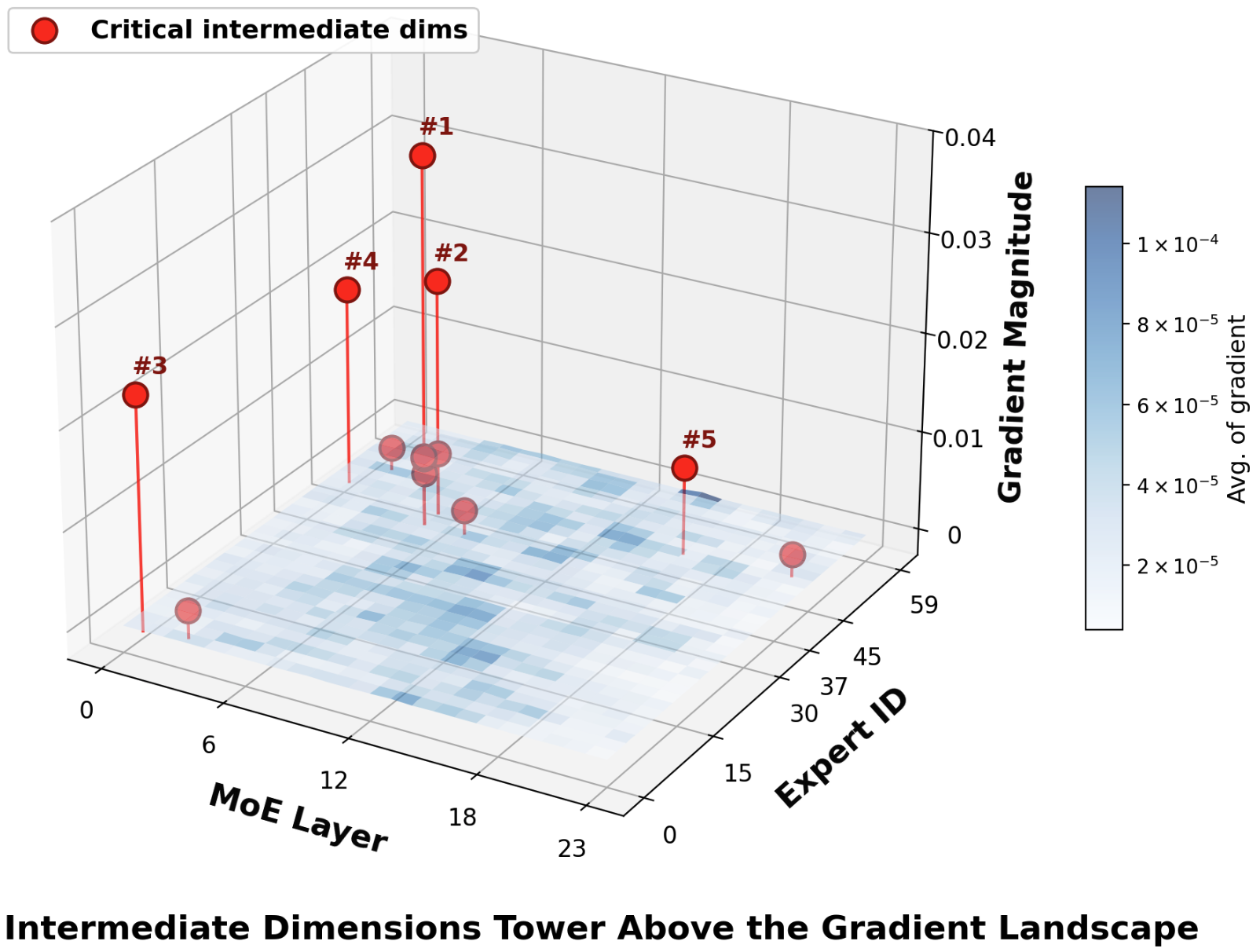}
\caption{Fisher scores of intermediate dimensions across experts and layers in Qwen1.5-MoE.}
\label{fig:critical_neurons_surface}
\end{figure}

We rank all $\sim$$1.35\text{M}$ MoE FFN intermediate dimensions by Fisher score (computed on $128$ \texttt{GSM8K} training samples) and zero-mask the top fraction.
Masking just $12$ of $1.35\text{M}$ intermediate dimensions ($0.001\%$) collapses \texttt{GSM8K} from $35.9\%$ to $0.8\%$ and \texttt{MATH} from $13.0\%$ to $1.1\%$, with code and \texttt{BBH} also degrading sharply, while multiple-choice knowledge tasks (\texttt{MMLU/CEval/CMMLU}) largely retain base (Table~\ref{tab:reverse_cross_domain}; full breakdown in Table~\ref{tab:raw_table1}). Appendix~\ref{appendix:gsm8k_failure_modes} confirms none of the residual $0.8\%$ on \texttt{GSM8K} reflects real computation -- all $10$ ``correct'' answers are coincidental digit matches. Figure~\ref{fig:critical_neurons_surface} shows the underlying cause: the Fisher distribution is extremely heavy-tailed, with these twelve dimensions $\sim$$1000\times$ above the population mean.

We further measure each dimension's mean forward activation magnitude. The population is sharply skewed: the top-12 Fisher dimensions carry mean $7$--$450\times$ the median activation of all parameters. (Appendix~\ref{sec:massive_activation}) They match the \emph{massive-activation} signature of~\citet{sun2024massive}, further suggest Fisher identify critical parameters.

To understand why masking only twelve dimensions devastates math reasoning, we trace the failure into the model's attention dynamics. 
\emph{Attention sink}~\citep{xiao2023streamingllm} stabilises autoregressive decoding. Masking the top-12 collapses mean \texttt{<BOS>} attention in the mid-stack (layer 7$\sim$17, where the sink dominates) by $\sim$$70\%$ reduction (Appendix~\ref{sec:attention_sink}). This directly explains the dissociation in Table~\ref{tab:reverse_cross_domain}: multi-step generation depends on the sink-stabilised token-to-token dynamics and degenerates into the echo and garbled outputs of Appendix~\ref{appendix:gsm8k_failure_modes}, whereas MCQ tasks need only a single-token classification and are largely sink-insensitive. 

The cross-domain analysis from \S\ref{sec:intermediate-dimension-overlap} suggests why removing \texttt{GSM8K}-calibrated critical dimensions selectively destroys math reasoning. Table~\ref{tab:pairwise-intermedia-dimension-overlap} shows that the pairwise overlap of Fisher-important dimensions between tasks exhibits clear domain structure: math tasks (\texttt{GSM8K/MATH}) share 65.3\% of their important dimensions. The high overlap explains why removing \texttt{GSM8K}-calibrated critical dimensions also destroys \texttt{MATH} performance, while the lower overlap with knowledge tasks (${\sim}$55\%) suggests why \texttt{MMLU/CEval/CMMLU} remain intact.

\subsection{Identify Universally Indispensable Dimensions}
\label{subsec:shared_dimensions}

The domain-overlap analysis identifies a set of intermediate dimensions retained by all eight evaluation tasks under $p=50\%$ ($\mathcal{K}_{\cap}$, 4.88\% of parameters). Removing the universally \emph{kept} dimensions collapses every task to near-zero accuracy. 
Removing the universally \emph{discarded} dimensions ($\mathcal{D}_{\cap}$, 4.01\% of parameters) preserves near-full performance across all tasks (Table~\ref{tab:reverse_cross_domain}), confirming these dimensions are genuinely redundant. This result, combined with \S\ref{sec:critical_dimensions}, leads to a key insight for compression: removing the long redundant tail of intermediate dimensions ranked by Fisher preserves model performance. We leverage this in \S\ref{sec:intermediate_compression}.

\section{Keep the Experts, Slim Their FFNs}
\label{sec:intermediate_compression}

Section~\ref{sec:gradient_metrics} showed that Fisher ranks parameters in a way that aligns with downstream impact: removing the top-ranked most disrupts generation, while the bottom-ranked form a long redundant tail. In this section, we leverage this ranking for compression. We first formalize prior expert-level compression and our finer-grained intermediate dimension alternative (\S\ref{subsec:compression_formulation}), then demonstrate that intermediate dimension compression substantially outperforms expert-level methods (\S\ref{subsec:intermediate_vs_expert}), and finally quantify the deployment benefits (\S\ref{subsec:deployment_cost}).

\subsection{From Expert-Level to Intermediate Dimension Compression}
\label{subsec:compression_formulation}

A Mixture-of-Experts layer consists of $n$ expert combined through a learned routing mechanism. Given an input $\mathbf{x} \in \mathbb{R}^{d}$, the MoE layer output is
\begin{equation}
f(\mathbf{x};\Phi) = \sum_{i \in T(\mathbf{x})} g_i(\mathbf{x})\, E_i(\mathbf{x};\theta_i),
\end{equation}
where $T(\mathbf{x})$ denotes the set of top-$k$ experts selected by the router, $g_i(\mathbf{x})$ is the routing weight for expert $i$, and $\Phi = \{(\theta_i, W_i, b_i)\}_{i=1}^n$ collects all layer parameters.
Let $\mathcal{P}_{\text{MoE}}$ be all of the experts FFN weights, and $\hat{\mathcal{P}}_{\text{MoE}}$ what remains after compression. We define the \textbf{\textit{MoE compression ratio}} as
$p \;=\; 1 - \frac{|\hat{\mathcal{P}}_{\text{MoE}}|}{|\mathcal{P}_{\text{MoE}}|}.
$ At $p=50\%$, expert-level methods remove half of the experts, and intermediate dimension compression reduces $d_{\text{ff}}$ inside experts. Because the MLP weights dominate LLM parameter counts, while attention, embeddings, router, and layer norms are never compressed, $p=50\%$ still translates to a  43--48\% whole-model compression rate $p_{\text{model}}$ across backbones (reported in Table~\ref{tab:model_size}).

\paragraph{Expert-level compression.}
Prior methods select a subset $\mathcal{S} \subset \{1,\dots,n\}$ of experts to remove in each MoE layer. The compressed layer becomes
\begin{equation}
\hat{f}(\mathbf{x}) = \sum_{i \in T(\mathbf{x}) \setminus \mathcal{S}} \hat{g}_i(\mathbf{x})\, E_i(\mathbf{x};\theta_i),
\end{equation}
where $\hat{g}_i$ denotes re-normalized routing weights. Expert selection uses activation ratio, router score, weight magnitude, or Fisher.

\begin{figure}[t]
\centering
\includegraphics[width=\columnwidth]{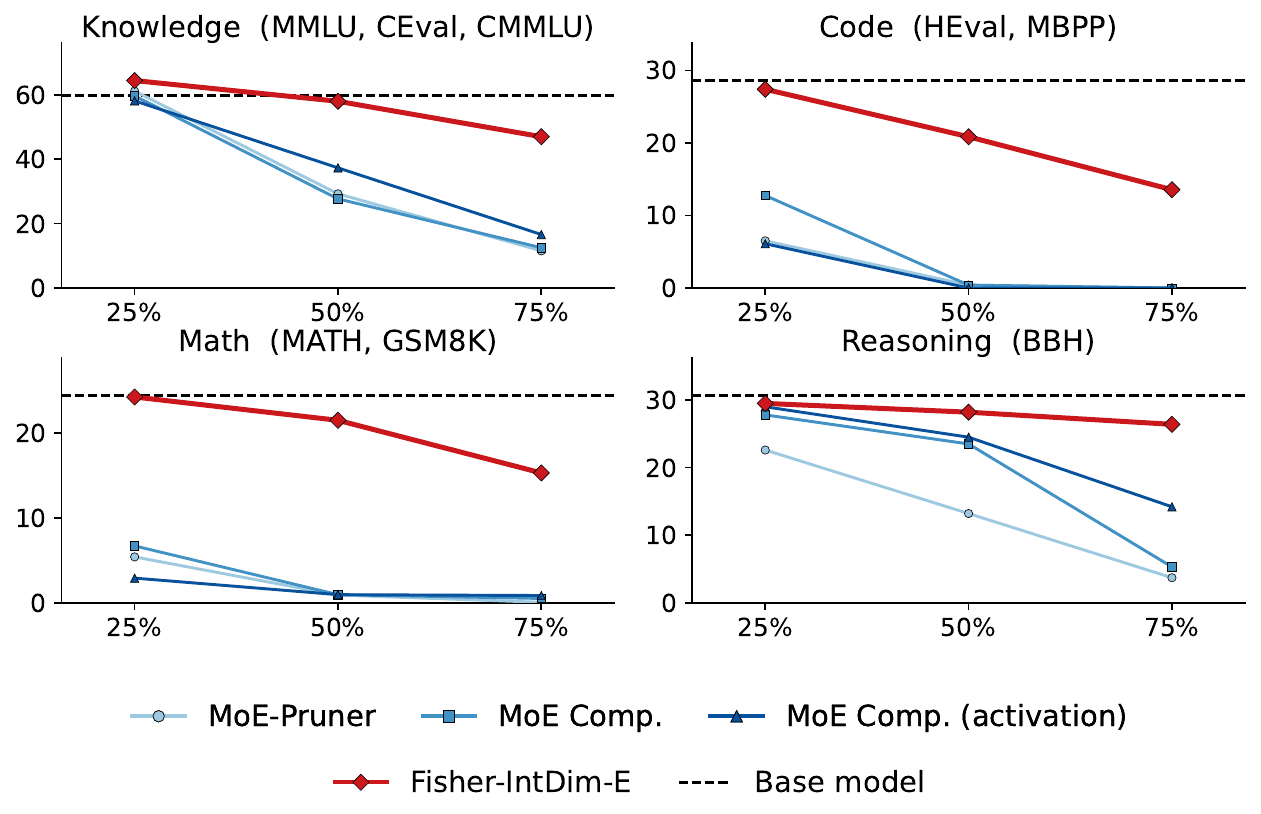}
\caption{Performance of Qwen1.5-MoE under expert-level compression at three drop ratios (25\%, 50\%, 75\%).}
\label{fig:moe-prune-by-ratio}
\end{figure}

\paragraph{Proposed Intermediate dimension compression.}
Instead of removing entire experts, we operate at the granularity of individual FFN intermediate dimensions. For each expert $E_i$ in each MoE layer, we use the Fisher importance score $s_{i,j}^{\text{dim}}$ (Eq.~\ref{eq:dim_importance}, formulated in \S\ref{subsec:formalize_metrics}) to rank every intermediate dimension $j \in \{1, \dots, d_{\text{ff}}\}$ and remove the lowest-scoring fraction $p$. 
The compressed expert becomes
\begin{equation}
\hat{E}_i(\mathbf{x}) = \hat{W}_i^{\text{down}} \left(\sigma(\hat{W}_i^{\text{gate}} \mathbf{x}) \odot \hat{W}_i^{\text{up}} \mathbf{x}\right),
\label{eq:compressed_expert}
\end{equation}
where $\hat{W}_i^{\text{gate}}, \hat{W}_i^{\text{up}} \in \mathbb{R}^{\hat{d}_{\text{ff}} \times d}$ and $\hat{W}_i^{\text{down}} \in \mathbb{R}^{d \times \hat{d}_{\text{ff}}}$ are obtained by physically removing the rows of $W^{\text{gate}}_i, W^{\text{up}}_i$ and the corresponding columns of $W^{\text{down}}_i$
. The removal is determined by the Fisher $s_{i,j}^{\text{dim}}$ together with a budget rule that fixes \emph{where} the budget $p$ is spent. We instantiate three rules: \textbf{IntDim-E} keeps the top $(1{-}p)$ fraction of dimensions within each expert; \textbf{IntDim-L} pools dimensions across experts in a layer and keeps the top $(1{-}p)$ fraction per layer; \textbf{IntDim-G} pools dimensions across the whole model and keeps the top $(1{-}p)$ globally. All three share the same score and budget to isolate the effect of allocation flexibility.
They reduce the actual parameter count and computation per expert and
preserve the model's routing behavior. We choose Fisher importance as our scoring criterion, as justified in \S\ref{sec:gradient_metrics}.

\subsection{Intermediate Dimension Compression Outperforms Expert-Level Methods}
\label{subsec:intermediate_vs_expert}


\begin{figure}[t]
\centering
\includegraphics[width=\columnwidth]{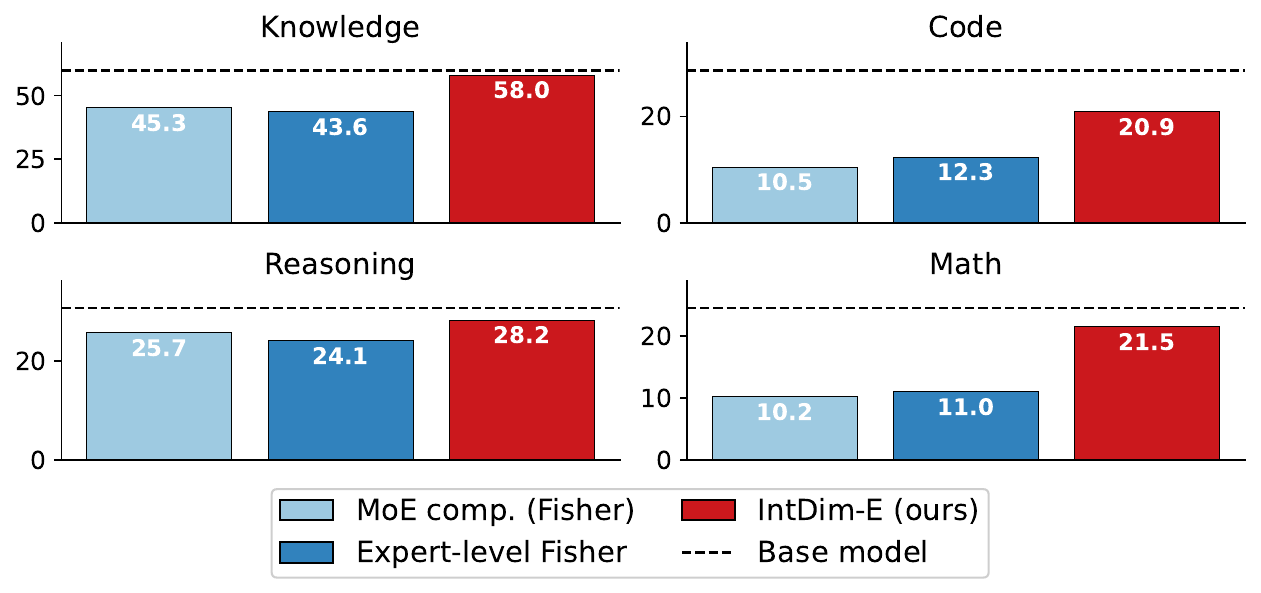}
\caption{Expert-level vs.\ intermediate dimension compression at $p=50\%$.}
\label{tab:expert-vs-intermediate}
\end{figure}
We compare expert-level and intermediate dimension compression at the same compression ratio $p=50\%$ on Qwen1.5-MoE-A2.7B across eight general-purpose benchmarks. Results on additional backbones, including OLMoE, Qwen3-30B, and Qwen3.5-35B, are provided in~\S\ref{sec:experiment}. Both use Fisher and domain-specific calibration data (\texttt{GSM8K} 128 training samples). At $p=50\%$, expert-level compression removes half routed experts per layer; IntDim-E reduces $d_{\text{ff}}$ by half within each expert.


Table~\ref{tab:expert-vs-intermediate} (Full results in Table~\ref{tab:raw_table2}) reveals that intermediate dimension compression dramatically outperforms expert-level compression including Expert-level Fisher across all benchmarks. We further scale our experiments across different compression ratios in Fig.~\ref{fig:moe-prune-by-ratio} (Full results in Table~\ref{tab:raw_fig3}): with $p = 25\%, 50\%, 75\%$, intermediate dimension removal consistently retains base model performance and outperform expert-level methods. In particular, comparisons with Expert-level Fisher highlight the advantage of fine-grained compression.



The core reason for this advantage is that knowledge in MoE models is distributed across all experts in a long-tailed manner, rather than concentrated in a few~\citep{he2026preserving}. Each expert contains both essential and redundant intermediate dimensions. Expert-level compression must discard entire experts including their essential components. Whereas intermediate dimension compression selectively removes only the least important computation within each expert, precisely the dimensions that \S\ref{subsec:shared_dimensions} showed are genuinely dispensable.

\subsection{Deployment Cost and Inference Speedup}
\label{subsec:deployment_cost}

\begin{table}[h]
\centering
\small
\setlength{\tabcolsep}{4pt}
\begin{tabular}{lccc}
\toprule
\textbf{Metric} & \textbf{Base} & \textbf{Compressed} & \textbf{Change} \\
\midrule
Latency $\downarrow$
& 2.233\,s & 1.839\,s & $1.21\times \downarrow$ \\
Throughput $\uparrow$
& 7,336\,tok/s & 8,909\,tok/s & $1.21\times \uparrow$ \\
KV-cache budget $\uparrow$
& 187K & 251K & $+33.7\%$ \\
VRAM $\downarrow$
& 26.67\,GiB & 15.09\,GiB & $-43.4\%$ \\
Active params $\downarrow$
& 2.689\,B & 2.274\,B & $-15.4\%$ \\
\bottomrule
\end{tabular}
\caption{Inference efficiency of 50\% compressed Qwen1.5-MoE on H100.}
\label{tab:deployment_inference_efficiency}
\end{table}

Intermediate dimension compression achieves \textbf{\textit{(1) parameter reduction}} compared to the base model: the gate, up, and down projection matrices are physically resized,
where $\hat{W}_i^{\text{gate}}, \hat{W}_i^{\text{up}} \in \mathbb{R}^{\hat{d}_{\text{ff}} \times d}, \hat{W}_i^{\text{down}} \in \mathbb{R}^{d \times \hat{d}_{\text{ff}}}$ are compressed from $W^{\text{gate}}_i, W^{\text{up}}_i \in \mathbb{R}^{d_{\text{ff}} \times d}, W^{\text{down}}_i \in \mathbb{R}^{d \times d_{\text{ff}}}$,
producing a smaller model.
\textbf{\textit{(2) Inference speedup}} compared to the base model and other expert-level removal baselines: Expert removal methods~\citep{muzio2024seer, lu-etal-2024-experts, chen2022task, li2023merge, chen2025retraining} discard entire experts from MoE layers but keep the number of activation parameters per token the same. The FFN intermediate dimension reduction translates to reduced active-parameter savings per token instead. 

We benchmark the 50\% compress ratio Qwen1.5-MoE against the base model on a single NVIDIA H100 GPU using vLLM with bf16 precision.\footnote{We greedy decode 1024 prompts and \textit{max\_new\_tokens} = 256. We use 1024-prompt warm-up precedes each timed run and \texttt{gpu\_memory\_utilization=0.85}.} 
Concretely, each layer goes from $86.1$M to $68.8$M parameters (a 20\% per-layer reduction). The theoretical decode-time speedup for Qwen1.5-MoE is
\texttt{$2.689\text{B} / 2.274\text{B} \approx 1.18\times.$}
Table~\ref{tab:deployment_inference_efficiency} shows that the IntDim-E compressed model achieves a $1.21\times$ wall-clock speedup, slightly exceeding the theoretical estimate. The additional gain comes from the freed VRAM being automatically converted by vLLM into a larger KV cache, increasing the request capacity by 34\%. The VRAM and disk footprint reduces from 27\,GB to 16\,GB. 

\section{Experiments}
\label{sec:experiment}
\begin{table*}[t]
\centering
\setlength{\tabcolsep}{2pt}
\renewcommand{\arraystretch}{1.2}
\caption{Performance of different pruning strategies at MoE compression ratio $p=50\%$. \textbf{Upper:} Qwen1.5-MoE-A2.7B and OLMoE-1B-7B on general benchmarks. \textbf{Lower:} Qwen3-30B-A3B and Qwen3.5-35B-A3B on frontier long-CoT math reasoning. Calibration uses the corresponding domain training data. Here, ``HE'' is short for HumanEval, ``MA'' is short for MultiArith, and ``Reas.'' is short for Reasoning.}
\label{tab:pruning-general}
\label{tab:pruning-reasoning}

\resizebox{\linewidth}{!}{%
\begin{tabular}{l
  | >{\columncolor{blue!2}}c >{\columncolor{blue!2}}c >{\columncolor{blue!2}}c
  | >{\columncolor{blue!6}}c >{\columncolor{blue!6}}c
  | >{\columncolor{blue!10}}c >{\columncolor{blue!10}}c >{\columncolor{blue!10}}c
  | >{\columncolor{blue!14}}c
  | >{\columncolor{red!12}}c
  | >{\columncolor{blue!2}}c >{\columncolor{blue!2}}c >{\columncolor{blue!2}}c
  | >{\columncolor{blue!6}}c >{\columncolor{blue!6}}c
  | >{\columncolor{blue!10}}c >{\columncolor{blue!10}}c >{\columncolor{blue!10}}c
  | >{\columncolor{blue!14}}c
  | >{\columncolor{red!12}}c}
\hline
\multicolumn{1}{c|}{\multirow{3}{*}{\textbf{Method}}}
& \multicolumn{10}{c|}{\textit{\textbf{Qwen1.5-MoE-A2.7B}}}
& \multicolumn{10}{c}{\textit{\textbf{OLMoE-1B-7B}}} \\
\cline{2-21}
& \multicolumn{3}{c|}{\cellcolor{blue!2}\textbf{Knowledge}}
& \multicolumn{2}{c|}{\cellcolor{blue!6}\textbf{Coding}}
& \multicolumn{3}{c|}{\cellcolor{blue!10}\textbf{Math}}
& \multicolumn{1}{c|}{\cellcolor{blue!14}\textbf{Reas.}}
& \multicolumn{1}{c|}{\cellcolor{red!12}\textbf{Avg.}}
& \multicolumn{3}{c|}{\cellcolor{blue!2}\textbf{Knowledge}}
& \multicolumn{2}{c|}{\cellcolor{blue!6}\textbf{Coding}}
& \multicolumn{3}{c|}{\cellcolor{blue!10}\textbf{Math}}
& \multicolumn{1}{c|}{\cellcolor{blue!14}\textbf{Reas.}}
& \multicolumn{1}{c}{\cellcolor{red!12}\textbf{Avg.}} \\
\cline{2-21}
& \textbf{MMLU} & \textbf{CEval} & \textbf{CMMLU} & \textbf{HE} & \textbf{MBPP} & \textbf{MATH} & \textbf{GSM8K} & \textbf{MR} & \textbf{BBH} &  
& \textbf{MMLU} & \textbf{CEval} & \textbf{CMMLU} & \textbf{HE} & \textbf{MBPP} & \textbf{MATH} & \textbf{GSM8K} & \textbf{MA} & \textbf{BBH} &  \\
\hline
Base Model                      & 59.3 & 59.3 & 60.7 & 32.6 & 24.6 & 13.0 & 35.9 & 76.7 & 30.7 & 43.3 & 53.5 & 33.1 & 33.0 &  9.8 & 15.2 & 11.6 & 53.2 & 87.5 & 26.5 & 35.9 \\
\hline
MoE comp. (activation)          & 26.3 & 42.5 & 43.1 &  0.0 &  0.0 &  0.0 &  1.9 &  3.3 & 24.5 & 15.7 & 33.2 & 22.5 & 26.0 &  9.2 & 11.9 &  1.5 &  5.0 & 22.2 & 22.6 & 17.1 \\
MoE comp.                       & 37.2 & 13.5 & 32.3 &  0.0 &  0.8 &  0.0 &  1.9 &  2.0 & 23.5 & 12.4 & 31.6 & 22.6 & 27.1 &  9.2 & 10.6 &  3.5 &  7.1 & 17.3 & 24.7 & 17.1 \\
MoE-Pruner\tablefootnote{For OLMoE-1B-7B, MoE-Pruner uses low-magnitude weights as the importance metric, since the original high-magnitude variant collapses accuracy across all tasks.}
                                & 18.6 & 33.3 & 35.8 &  0.8 &  0.0 &  0.1 &  1.7 &  2.7 & 13.2 & 11.8 & 17.3 & 12.9 & 19.0 &  0.0 &  0.0 &  0.0 &  2.1 &  2.7 & 12.2 &  7.4 \\
MoE comp. (Fisher)              & 39.4 & 48.8 & 47.7 & 12.9 &  8.1 &  2.0 & 16.9 & 15.5 & 25.7 & 24.1 & 33.8 & 27.0 & 29.0 &  5.5 &  9.1 &  1.9 &  5.2 & 26.5 & 24.0 & 18.0 \\
Expert-level Fisher             & 37.9 & 48.6 & 44.2 & 12.9 & 11.7 &  2.4 & 18.0 & 37.7 & 24.1 & 26.4 & 34.4 & 27.7 & 27.6 &  7.3 & 10.1 &  1.7 &  6.1 & 21.5 & 25.4 & 18.0 \\
\textbf{Fisher-IntDim-E (ours)} & \hK 50.3 & \hK 62.0 & 61.6 & 21.2 & 20.5 & 8.0 & \hM 35.0 & 61.5 & \hR 28.2 & \hA 38.7 & \hK 39.9 & \hK 29.2 & \hK 29.9 & \hC 9.8 & 7.3 & 2.3 & 24.9 & 75.7 & \hR 27.3 & 27.4 \\
\textbf{Fisher-IntDim-L (ours)} & \hK 49.1 & 57.5 & 62.0 & 30.5 & 22.8 & \hM 12.6 & 22.0 & \hM 89.8 & \hR 28.3 & \hA 41.6 & \hK 40.4 & \hK 30.8 & \hK 31.4 & \hC 9.8 & \hC 17.0 & \hM 7.8 & \hM 44.3 & \hM 87.7 & \hR 25.1 & \hA 32.7 \\
\textbf{Fisher-IntDim-G (ours) \space \space} & \hK 49.0 & \hK 61.3 & \hK 64.9 & \hC 34.2 & \hC 24.1 & \hM 9.8 & \hM 27.2 & \hM 92.2 & \hR 28.2 & \hA 43.4 & \hK 40.4 & \hK 31.4 & \hK 31.7 & \hC 10.4 & \hC 15.7 & \hM 6.9 & \hM 41.9 & \hM 81.0 & \hR 25.2 & \hA 31.6 \\
\hline
\end{tabular}}

\vspace{2pt}

\resizebox{\linewidth}{!}{%
\begin{tabular}{l 
  | >{\columncolor{blue!10}}c >{\columncolor{blue!10}}c >{\columncolor{blue!10}}c >{\columncolor{blue!10}}c >{\columncolor{blue!10}}c
  | >{\columncolor{blue!14}}c
  | >{\columncolor{red!12}}c
  | >{\columncolor{blue!10}}c >{\columncolor{blue!10}}c >{\columncolor{blue!10}}c >{\columncolor{blue!10}}c >{\columncolor{blue!10}}c
  | >{\columncolor{blue!14}}c
  | >{\columncolor{red!12}}c
  }
\hline
\multicolumn{1}{c|}{\multirow{3}{*}{\textbf{Method}}}
& \multicolumn{7}{c|}{\textit{\textbf{Qwen3-30B-A3B}}}
& \multicolumn{7}{c}{\textit{\textbf{Qwen3.5-35B-A3B}}} \\
\cline{2-15}
& \textbf{AIME25} & \textbf{AIME26} & \textbf{Olympiad} & \textbf{MATH-500} & \textbf{GSM8K} & \textbf{GPQA-D} & \textbf{Avg.}
& \textbf{AIME25} & \textbf{AIME26} & \textbf{Olympiad} & \textbf{MATH-500} & \textbf{GSM8K} & \textbf{GPQA-D} & \textbf{Avg.} \\
\hline
Base Model              & 33.3 & 50.0 & 49.6 & 64.2 & 61.9 & 48.5 & 51.3 & 66.7 & 66.7 & 77.9 & 92.4 & 48.8 & 72.2 & 70.8 \\
MoE-Pruner              &  0.0 &  0.0 &  0.3 &  2.4 &  0.8 &  5.1 &  1.4 & 13.3 & 10.0 & 22.1 & 34.8 & 17.4 & 19.7 & 19.6 \\
MoE comp.               &  0.0 &  0.0 &  0.9 &  1.8 &  2.3 & 10.6 &  2.6 &  0.0 &  0.0 &  0.0 &  0.2 &  2.1 & 12.6 &  2.5 \\
MoE comp. (activation)  &  0.0 &  0.0 &  1.2 &  1.6 &  2.8 &  2.5 &  1.4 &  0.0 &  0.0 &  0.2 &  0.2 &  2.3 &  9.6 &  2.0 \\
MoE comp. (Fisher)      & 46.7 & 50.0 & 55.0 & \hM 76.4 & \hM 92.3 & 38.9 & 59.9 & \hM 80.0 & \hM 76.7 & \hM 79.8 & \hM 91.8 & 42.2 & \hR 71.7 & 73.7 \\
Expert-level Fisher     & 36.7 & 53.3 & 54.9 & \hM 76.2 & \hM 93.4 & \hR 49.0 & 60.6 & \hM 80.0 & \hM 76.7 & \hM 81.9 & \hM 92.4 & 50.7 & \hR 69.7 &  75.2 \\
\textbf{Fisher-IntDim-G (ours)} & \hM 60.0 & \hM 56.7 & \hM 58.6 & \hM 74.6 & 81.4 & \hR 50.0 & \hA 63.5 & \hM 76.7 & \hM 73.3 & \hM 79.5 & \hM 92.4 & \hM 62.6 & \hR 70.2 & \hA 75.8 \\
\hline
\end{tabular}}
\end{table*}

In Subsection~\S\ref{subsec:exp_general}, we study how different MoE compression strategies behave when directly applied to the base model without any post-training. We compress Qwen1.5-MoE-A2.7B and OLMoE-1B-7B-0125 at $p=50\%$ under task-matched domain calibration, and report performance across the eight general-task benchmarks above to isolate the effect of the importance metric and compression granularity.
In Subsection~\S\ref{subsec:exp_math_reasoning}, we study whether the granularity advantage transfers to long-CoT mathematical reasoning at frontier scale on Qwen3-30B-A3B and Qwen3.5-35B-A3B with 128 \texttt{Stanford-S1} calibration data. In Subsection~\S\ref{subsec:exp_sft}, we study whether post-trained compression models can achieve competitive results compared to full post-trained models. 
In Subsection~\S\ref{subsec:exp_ood}, we study the out-of-domain generalization of single-domain calibration. We use \texttt{GSM8K} calibration for all methods and evaluate the resulting model on the seven non-math benchmarks to test whether the Fisher signal preserves general capability when the calibration distribution is narrow. In Subsection~\S\ref{subsec:exp_quant}, we empirically show that Fisher-MoE composes with quantization to further reduce model deployment cost. We apply 4-bit AWQ~\citep{lin2024awq} on top of Fisher-MoE at $p=50\%$ and compare the accuracy and footprint of \emph{Ours+AWQ} against \emph{Base+AWQ} to verify that the two reductions compose without amplifying quantization sensitivity. 

After evaluating different compression ratios in Fig.~\ref{fig:moe-prune-by-ratio} ($p = 25\%, 50\%, 75\%$), we fix the MoE compression ratio to $p = 50\%$ (defined in \S\ref{subsec:compression_formulation}, the \emph{whole-model} parameter reduction is $\sim$45\% across four backbones; see Table~\ref{tab:model_size}) for the remainder of this section, and focus on scaling across models and benchmarks, along with generalization and compatibility. Detailed experimental settings, including models and dataset, baselines, calibration data, computational resources, and training and evaluation framework are in Appendix~\ref{appendix:appendix_experiment_settings}. We include an ablation study and cost analysis on the calibration sample size $N \in {32, 64, 128, 256, 512}$ in Appendix~\ref{appendix:calibrarion_ablation_study}, and use $N=128$; all methods use the same calibration data.
128 samples cost less than 30 seconds for Fisher calculation on an H100 node for the 30B models. Further comparisons with dense pruning and sparsity baselines are in Appendix~\ref{app:dense-pruning}, though they fall outside the scope of MoE architectures.

\subsection{Compression on Base Models: General Tasks}
\label{subsec:exp_general}

We compress each base model at the MoE compression ratio $p=50\%$ and evaluate across the eight general-purpose benchmarks. This isolates the effect of the compression algorithm: any performance loss reflects information that the importance metric and granularity failed to preserve.

Table~\ref{tab:pruning-general} shows a consistent picture across both backbones. Expert-level baselines, regardless of importance metric, suffer catastrophic collapse on generation-heavy tasks, including \texttt{GSM8K}, \texttt{MATH}, and \texttt{HumanEval}. Fisher-MoE, on the other hand, preserves generation capability: it largely retains capability in math and code reasoning datasets. The advantage holds across knowledge tasks too. 

Additionally, enabling more flexible allocation of intermediate dimension leads to further gains. Specifically, Fisher-IntDim-G outperforms Fisher-IntDim-E on both OLMoE and Qwen1.5-MoE. In the remainder of this section, we adopt Fisher-IntDim-G for scaling experiments, including larger models, long chain-of-thought (CoT) math reasoning, and Supervised Fine-tuning (SFT).



On \texttt{CEval}, \texttt{CMMLU}, and \texttt{MultiArith}, Fisher-MoE matches or modestly exceeds the uncompressed base model. 
We interpret these gains as evidence that Fisher-based pruning can sometimes remove dimensions associated with brittle shortcut behavior or out-of-distribution output formats, thereby revealing capabilities already present in the base model rather than introducing new ones. 
A similar pattern is observed for the Qwen3 and Qwen3.5 models in \S\ref{subsec:exp_math_reasoning}. 
We further analyze this phenomenon quantitatively in Appendix~\ref{appendix:mechanisms}. 
Compression reduces generation-time biases that suppress useful computation already available in the original model. 
Finally, we show that removing entire experts or attention heads causes severe degradation on generation-heavy benchmarks, while FFN intermediate dimension compression is substantially more robust (Appendix~\ref{appendix:redundancy_locus}).

\subsection{Compression on Base Models: Math Reasoning at Larger Scale}
\label{subsec:exp_math_reasoning}

We next test whether the granularity advantage carries over to long-CoT math reasoning at larger scale, where errors compound across many decoding steps and any loss of generation capability is unforgiving. We compress Qwen3-30B-A3B and Qwen3.5-35B-A3B at the same MoE compression ratio $p=50\%$ and evaluate on five long-CoT reasoning benchmarks in Tab.~\ref{tab:pruning-reasoning}.

\subsection{Fine-Tuning Enabled Domain Enhancement}
\label{subsec:exp_sft}

The base-model results above measure the naive case: compressed weights deployed without any further training. In practice, a compressed model is often used as the starting point for domain SFT. The relevant question for deployment is therefore: \emph{does compressing first then SFTing recover the base + SFT ceiling?} We compress at $p=50\%$, then SFT on the corresponding domain training data, and compare against the uncompressed base + SFT.

\begin{table*}[t]
\centering
\begin{minipage}[t]{0.487\textwidth}
\centering
\textit{(a) Qwen1.5-MoE-A2.7B}\\[2pt]
\resizebox{\textwidth}{!}{%
\begin{tabular}{lccc}
\toprule
\textbf{Method} & \textbf{MATH} & \textbf{GSM8K} & \textbf{MultiArith} \\
\midrule
Base + SFT                                            & 15.9 & 50.8 & 87.8 \\
\midrule
MoE-Pruner + SFT                                      &  3.7 & 40.6 & 35.8 \\
MoE comp. + SFT                                 &  4.9 & 30.3 & 44.7 \\
MoE comp. (activation) + SFT                    &  4.8 & 28.5 & 47.7 \\
MoE comp. (Fisher) + SFT     & 14.3 & 46.0 & 81.8 \\
Expert-level Fisher + SFT                                            & 15.4 & 46.4 & 84.8 \\
\rowcolor{blue!8}
\textbf{Fisher-IntDim-G + SFT (ours)}                        & 17.6 & 52.0 & 88.0 \\
\bottomrule
\end{tabular}}
\end{minipage}\hfill
\begin{minipage}[t]{0.49\textwidth}
\centering
\textit{(b) OLMoE-1B-7B-0125}\\[2pt]
\resizebox{\textwidth}{!}{%
\begin{tabular}{lccc}
\toprule
\textbf{Method} & \textbf{MATH} & \textbf{GSM8K} & \textbf{MultiArith} \\
\midrule
Base + SFT                                            & 12.4 & 45.1 & 92.5 \\
\midrule
MoE-Pruner + SFT                                      &  2.3 & 15.2 & 39.5 \\
MoE comp. + SFT                                 &  9.4 & 30.6 & 87.8 \\
MoE comp. (activation) + SFT                    &  9.8 & 31.8 & 87.0 \\
MoE comp. (Fisher) + SFT     & 11.4 & 37.1 & 89.3 \\
Expert-level Fisher + SFT                                            & 10.5 & 39.1 & 89.2 \\
\rowcolor{blue!8}
\textbf{Fisher-IntDim-G + SFT (ours)}                         & 11.9 & 43.8 & 92.3 \\
\bottomrule
\end{tabular}}
\end{minipage}
\caption{Compression at MoE ratio $p=50\%$ followed by SFT on domain training data. \emph{Base + SFT} is the uncompressed reference ceiling.}
\label{tab:pruning-sft}

\end{table*}

Results in Tab.~\ref{tab:pruning-sft} show that using half the FFN parameters, after SFT, Fisher-MoE reaches competitive results compared with the uncompressed Base + SFT ceiling on \texttt{GSM8K}, \texttt{MultiArith}, and \texttt{MATH}. In contrast, the activation-, score-, and magnitude-based compression methods significantly under-perform even after SFT.

\subsection{Out-of-Domain Generalization}
\label{subsec:exp_ood}

\begin{table}[h]
  \centering
  \caption{Evaluation of Qwen1.5-MoE compressed models. \texttt{GSM8K} is used for calibration. In-domain tasks include arithmetic reasoning (\texttt{MATH}/\texttt{GSM8K}/\texttt{MultiArith}); out-of-domain results are grouped into Knowledge (\texttt{MMLU/CEval/CMMLU}), Code (\texttt{HumanEval/MBPP}) and Reasoning (\texttt{BBH}). Full results in Table~\ref{tab:raw_table7}.}
  \label{tab:ood}
  \resizebox{\linewidth}{!}{%
  \begin{tabular}{l ccc|ccc}
  \toprule
  \textbf{Method} 
  & \multicolumn{3}{c|}{\textbf{In-domain (Math)}} 
  & \multicolumn{3}{c}{\textbf{Out-of-domain}} \\
  \cmidrule(lr){2-4} \cmidrule(lr){5-7}
  \textbf{Qwen1.5-MoE-A2.7B} & \textbf{GSM8K (Calib)} & \textbf{MATH} & \textbf{MultiArith}
   & \textbf{Knowledge} & \textbf{Code} & \textbf{Reasoning} \\
  \midrule
  MoE-Pruner                              & 1.7  & 0.0 & 1.8 & 29.2 & 0.3 & 13.2 \\
  MoE comp.                               & 1.9  & 0.3 & 0.3 & 37.0 & 0.6 & 22.1 \\
  MoE comp. (activation)                  & 1.9  & 0.2 & 0.8 & 38.1 & 0.2 & 12.3 \\
  MoE comp. (Fisher)                      & 16.9 & 0.7 & 26.2 & 37.0 & 4.2 & 23.0 \\
  Expert-level Fisher                     & 18.0 & 0.7 & 31.7 & 38.7 & \textbf{5.8} & \textbf{24.6} \\
  \rowcolor{blue!8}
  \textbf{Fisher-IntDim-E (ours)}
                                          & \textbf{35.0} & \textbf{4.7} & \textbf{70.8}
                                          & \textbf{49.4} & \textbf{5.6} & \textbf{24.2} \\
  \bottomrule
  \end{tabular}%
  }
\end{table}

A reasonable concern with Fisher importance is that calibration on one domain may bias the retained dimensions toward that domain, hollowing out general capability. We test this directly: we calibrate on math (\texttt{GSM8K}, 128 samples), apply 50\% intermediate dimension compression, and evaluate the resulting model on seven Out-of-Domain (OOD) task categories. Then we further compare the OOD performance of 50\% intermediate dimension compression with other baselines. Results in Tab.~\ref{tab:ood} show that for both in-domain and OOD tasks, Fisher-IntDim-E dominates the activation-, score-, and magnitude-based baselines by large margins.

\subsection{Compatibility with Quantization}
\label{subsec:exp_quant}

Fisher-MoE reduces parameter count structurally; AWQ~\citep{lin2024awq} reduces bit-width per parameter. The two operate on orthogonal axes of model footprint, so combining them should multiply the savings if neither destroys the capability the other depends on. We test this by applying AWQ (settings in Appendix~\S\ref{appendix:awq_settings}) on top of IntDim-E at $p=50\%$ and against AWQ on the base model.

\begin{table}[h]
\centering
\caption{Stacking AWQ on top of Fisher-MoE at MoE ratio $p=50\%$. Full results in Table~\ref{tab:raw_table8}.}
\label{tab:main_results}
\resizebox{\columnwidth}{!}{%
\begin{tabular}{l cc| cccc c}
\toprule
\textbf{Method} 
& \textbf{Disk} & \textbf{VRAM} 
& \textbf{Knowledge} & \textbf{Code} & \textbf{Reasoning} & \textbf{Math} & \textbf{Avg.} \\
\midrule                       
Base Model               
& 26.7 & 58.1 
& 59.8 & 28.6 & 30.7 & 41.9 & 40.3 \\           
Base Model + AWQ          
& 7.9 & 8.6  
& 50.2 & 12.8 & 22.0 & 11.6 & 24.2 \\           
IntDim-E          
& 15.1 & 30.1 
& 58.0 & 20.9 & 28.2 & 34.8 & 35.5 \\
\rowcolor{blue!8}
\textbf{IntDim-E + AWQ} 
& 5.0 & 6.5  
& 50.2 & 13.3 & 24.6 & 13.5 & 25.4 \\  
\bottomrule
\end{tabular}%
}
\end{table}

The two reductions compose: combining \textit{IntDim-E} at $p=50\%$ with 4-bit AWQ shrinks the deployed footprint of Qwen1.5-MoE-A2.7B from 26.67\,GiB (bf16, base) to roughly 1/8 of that, while preserving the capability profile reported in \S\ref{subsec:deployment_cost}. The accuracy loss from \emph{IntDim-E+AWQ} relative to \emph{IntDim-E} matches the loss of \emph{Base Model+AWQ} relative to \emph{Base Model}, confirming that Fisher-MoE is compatible with quantization.

\section{Conclusion}
\label{sec:conclusion}
We revisit MoE compression through the lens of two questions: \emph{which signal} should drive parameter selection, and \emph{at what granularity} should compression operate. For the first, we show that the empirical Fisher information outperforms activation-, router-, and magnitude-based heuristics across backbone and benchmark (\S\ref{sec:gradient_metrics}, 
\S\ref{subsec:exp_general}. \S\ref{subsec:exp_math_reasoning}). For the second, we show that pushing compression below the expert level to FFN intermediate dimensions changes the regime (\S\ref{sec:intermediate_compression}, \S\ref{subsec:exp_general}). Beyond speedup and memory efficiency advantages (\S\ref{subsec:deployment_cost}, \S\ref{subsec:exp_quant}), our Fisher-importance attribution analysis localizes a striking dissociation (\S\ref{sec:critical_dimensions}), suggesting that intermediate dimension Fisher is a useful tool for both compression (\S~\ref{sec:intermediate_compression}) and understanding where capabilities live inside models (\S~\ref{sec:critical_dimensions}~\ref{subsec:shared_dimensions}).

\clearpage
\newpage
\section*{Limitations}
\label{sec:limitations}
Due to limited computational resources, we did not extend the exploration of Fisher-MoE to 100 Billion+ parameters MoE backbones, nor did we run a full recovery study with longer post-training on larger and more diverse corpora to characterize how much of the residual gap to the uncompressed base model can be closed. We leave both directions as natural follow-ups for future work.





\bibliography{custom}

\clearpage
\newpage
\appendix

\section{Related Work}
\label{sec:related_work}

Prior MoE compression methods can be grouped by their \emph{importance metric} (how compression targets are selected, we discussed in~\S~\ref{sec:introduction}).

\paragraph{Importance metrics.}
Across these strategies, methods differ in how they select which experts or parameters to compress. \textbf{Activation-ratio-based} methods measure routing frequency~\citep{muzio2024seer, lu-etal-2024-experts, chen2022task}. \textbf{Router-score-based} methods use average gating weights~\citep{xie2024moe, gu2025delta}. \textbf{Magnitude-based} methods rank by weight norms~\citep{lee2024stun, yang-etal-2024-moe}. MoE-Pruner~\citep{xie2024moe} combines gated values with weight magnitudes. Our work demonstrates that \textbf{gradient-based} importance---which directly measures each parameter's contribution to the loss---substantially outperforms all of these alternatives (\S\ref{subsec:formalize_metrics}). Moreover, all prior MoE compression methods operate at the granularity of entire experts or expert blocks; our intermediate dimension compression (\S\ref{sec:intermediate_compression}) represents a strictly finer granularity that better preserves the distributed knowledge structure of MoE models.

\paragraph{Model Attribution}
Our Fisher-importance-based model attribution analysis relates to the broader literature on understanding which model components are responsible for specific capabilities.
The discussion about parameter importance can trace back to Optimal Brain Damage~\citep{lecun1989optimal} and Optimal Brain Surgeon~\citep{hassibi1993optimal}, which use second-order information (Hessian diagonals) or first-order Taylor approximations~\citep{molchanov2017pruning} to estimate the loss increase from weight removal. Recent work extends these ideas to attention heads~\citep{michel2019sixteen, voita2019analyzing} or attention matrices~\citep{he2025sparsematrixlargelanguage}.
Several studies have investigated where factual knowledge is stored in transformers. \citet{geva2021transformer} show that FFN layers function as key-value memories. \citet{meng2022locating} use causal tracing to locate factual associations in specific MLP layers.
Our Fisher-importance-based attribution further provides a finer-grained lens for model attribution and understanding parameter specialization than layer-structure-wise analysis alone.

\newpage

\section{Critical Dimensions Coincide with FFN Activation Outliers}
\label{sec:massive_activation}

Section~\ref{sec:critical_dimensions} establishes that zero-masking the top-12 Fisher-ranked intermediate dimensions of Qwen1.5-MoE collapses GSM8K from $35.9\%$ to $0.8\%$ while leaving multi-choice knowledge tasks largely intact. This appendix examines the forward-pass activation behaviour of those twelve dimensions and places them within the broader literature on activation outliers~\citep{sun2024massive}. We make a deliberately limited claim: the critical dimensions are extreme upper-tail activation outliers that share the qualitative signature of -- but are not at the canonical scale of -- the massive-activation phenomenon documented in dense LLMs.

\paragraph{Fisher and activation magnitude are not statistically independent.}
Before reporting the activation statistics we make explicit a known mathematical dependency that prevents interpreting them as an independent confirmation of Fisher. For the down-projection slab, the gradient with respect to weight $(W^{\text{down}}_i)_{k,j}$ factors as
\(
\partial \mathcal{L} / \partial (W^{\text{down}}_i)_{k,j} \;=\; (\partial \mathcal{L} / \partial h^{(\ell)}_{k}) \cdot a^{(i)}_{j},
\)
where $a^{(i)}_{j}$ is the post-activation along intermediate dimension $j$ of expert $i$ and $\partial \mathcal{L}/\partial h^{(\ell)}_{k}$ is the upstream gradient at the residual stream. Squaring and summing this term across $k$ -- the form that enters the down-projection contribution of Eq.~\ref{eq:dim_importance} -- therefore scales with $(a^{(i)}_{j})^2$. High-activation channels accumulate Fisher mass through this term by construction, so any positive correlation between Fisher rank and activation magnitude is partly expected, not a separate signal. The gate- and up-projection contributions in Eq.~\ref{eq:dim_importance} do not factor in the same way, so the coupling is partial rather than total, but it is strong enough that we treat the activation analysis below as \emph{descriptive} -- characterising the magnitude regime the top-12 occupy and connecting them to the activation-outlier literature -- rather than as an independent attribution channel.

\begin{table}[t]
\centering
\small
\setlength{\tabcolsep}{6pt}
\begin{tabular}{l rrr}
\toprule
Rank & \textbf{mean $|a|$} & \textbf{$\times$ pop. med.} & \textbf{$\times$ pop. $p_{99}$} \\
\midrule
\emph{median} & $0.17$ & $1\times$ & $0.15\times$ \\
\midrule
\#1  & 26.1 & $150\times$ & $22\times$ \\
\#2  & 77.9 & $450\times$ & $66\times$ \\
\#3  & 20.4 & $118\times$ & $17\times$ \\
\#4  & 26.5 & $153\times$ & $23\times$ \\
\#5  & 12.7 & $73\times$  & $11\times$ \\
\#6  & 5.9  & $34\times$  & $5\times$  \\
\#7  & 4.7  & $27\times$  & $4\times$  \\
\#8  & 20.2 & $117\times$ & $17\times$ \\
\#9  & 4.0  & $23\times$  & $3\times$  \\
\#10 & 2.7  & $16\times$  & $2\times$  \\
\#11 & 1.3  & $7\times$   & $1\times$  \\
\#12 & 1.4  & $8\times$   & $1\times$  \\
\bottomrule
\end{tabular}
\caption{Forward activation magnitudes of the top-12 Fisher-ranked intermediate dimensions in Qwen1.5-MoE on GSM8K.}
\label{tab:massive_act_pop}
\end{table}

\paragraph{Measurement.}
For every routed expert in Qwen1.5-MoE-A2.7B and every intermediate dimension inside it (in total $\sim$$2.03\text{M}$ dimensions across $24$ MoE layers and $60$ experts per layer), we record the mean absolute post-activation $|a_j|$ on the GSM8K calibration set ($128$ samples, the same set used to compute Fisher). We then compare the population distribution of $|a_j|$ against the values observed at the twelve dimensions whose Fisher score is largest.

\paragraph{Population baseline is sharply skewed.}
The empirical distribution of mean $|a|$ is heavy-tailed even before we look at any Fisher-selected outliers: median $0.17$ and $p_{99}=1.17$. In other words, fewer than $1\%$ of all FFN intermediate dimensions exceed mean activation magnitude $1.17$, and the typical dimension carries a forward signal an order of magnitude smaller.

Top-12 Fisher dimensions are extreme upper-tail outliers -- but smaller than canonical massive activations.
Table~\ref{tab:massive_act_pop} reports, for each of the twelve dimensions, the mean $|a|$ together with its ratio to the population median and to the population $p_{99}$. The top nine dimensions carry mean $|a|$ between $4.0$ and $77.9$, i.e., $23\times$ to $450\times$ the population median and $3\times$ to $66\times$ the population $p_{99}$. Even the weakest two of the twelve (ranks $11$ and $12$) sit at $7$--$8\times$ the median, comfortably above the $99$th percentile of the entire MoE FFN. For calibration against the original literature, \citet{sun2024massive} report \emph{massive activations} in dense LLMs whose magnitudes exceed the channel mean by roughly $10^{4}\times$ -- two to three orders of magnitude beyond what we observe here. We therefore describe the top-12 as \emph{extreme upper-tail activation outliers} that share the qualitative signature of the massive-activation phenomenon -- a tiny subset of hidden coordinates carrying disproportionate forward-pass mass -- without claiming they are massive activations in the strict numerical sense of the original definition. The MoE setting may also dilute the per-channel magnitude relative to a dense backbone, since each expert is activated only on a fraction of tokens.

The Spearman rank correlation between Fisher and mean $|a|$ over the twelve dimensions is $0.91$. Given the gradient--activation coupling noted above, this correlation is the expected direction and approximately the expected strength: it confirms that intermediate-dimension Fisher concentrates on high-activation channels, but does not establish a statistically independent attribution channel. Read this way, the coupling is also the reason intermediate-dimension Fisher is a useful tractable attribution: by inheriting weight from the activation-driven down-projection term, it inherits an inductive bias toward exactly the outlier channels that prior work on activation outliers and attention sinks has shown to be load-bearing (Appendix~\ref{sec:attention_sink}).

\paragraph{Internal structure within the top-12.}
Ranks $1$--$9$ are extreme outliers deep in the upper tail of the population distribution ($\geq 23\times$ the median), while ranks $10$--$12$ are still above $p_{99}$ but no longer of the same magnitude ($7$--$16\times$). This tapering is consistent with activation outliers being concentrated in only a handful of channels per model -- the first nine ranks carry the bulk of the abnormal forward-pass mass that, as Appendix~\ref{sec:attention_sink} shows, is associated with the mid-stack attention sink.

\newpage
\section{Masking Critical Dimensions Reduces the Mid-Stack Attention Sink}
\label{sec:attention_sink}

Appendix~\ref{sec:massive_activation} shows that the twelve Fisher-critical intermediate dimensions of Qwen1.5-MoE coincide with extreme upper-tail outliers in the FFN activation distribution -- a milder MoE analogue of the massive activations documented in dense LLMs by~\citet{sun2024massive}. This appendix reports a second, downstream property of those dimensions: masking them substantially reduces the BOS attention sink~\citep{xiao2023streamingllm} in the mid-stack layers where the sink dominates. We use this observation to relate the accuracy dissociation in Table~\ref{tab:reverse_cross_domain} to a known mechanistic concept, while being explicit about what the experiment can and cannot establish.

In a softmax attention layer the attention weights for each query must sum to one. When no key is genuinely relevant, optimisation pressure has been shown to push probability mass onto a low-information ``parking'' position -- typically the BOS token~\citep{xiao2023streamingllm}. This sink stabilises decoding in two ways: it absorbs unallocated probability mass without distorting on-topic tokens, and it provides a residual-stream anchor whose norm keeps mid-stack pre-softmax logits inside a numerically well-behaved range. \citet{sun2024massive} further argue that the sink is supported by extreme FFN activations that progressively inflate the BOS residual-stream norm across depth.

We measure, per decoder layer, the mean fraction of softmax attention paid to the BOS token, averaged across attention heads and across all positions of every GSM8K test prompt. We compare the base model against the Mask-Top-12 variant -- the identical model with those twelve intermediate dimensions zeroed in the forward pass. Layers are grouped into bands that reflect the baseline sink profile of Qwen1.5-MoE.

\begin{table}[h]
\centering
\small
\setlength{\tabcolsep}{4pt}
\resizebox{\linewidth}{!}{%
\begin{tabular}{l ccc}
\toprule
\textbf{Layer band} & \textbf{base} & \textbf{Mask-Top-12} & \textbf{drop} \\
\midrule
L0--1             & $0.07$--$0.09$ & $0.07$--$0.09$ & $-0.00$ \\
L2--6              & $0.30$--$0.38$ & $0.22$--$0.35$ & $-0.05$ \\
\rowcolor{blue!8}
\textbf{L7--13}  & $\mathbf{0.34}$--$\mathbf{0.44}$ & $\mathbf{0.10}$--$\mathbf{0.18}$ & $\mathbf{-0.24}$ \\
\rowcolor{blue!8}
\textbf{L14--17}& $0.32$--$0.33$ & $0.07$--$0.10$ & $\mathbf{-0.25}$ \\
L18--21                     & $0.17$--$0.22$ & $0.02$--$0.03$ & $-0.17$ \\
L22--23              & $0.09$--$0.11$ & $0.03$--$0.04$ & $-0.06$ \\
\bottomrule
\end{tabular}}
\caption{Mean attention to BOS per decoder-layer band on GSM8K with Qwen1.5-MoE-A2.7B.}
\label{tab:sink_summary}
\end{table}

Masking the top-12 substantially reduces the mid-stack sink.
Table~\ref{tab:sink_summary} shows that Mask-Top-12 attenuates the sink in the layers where it is strongest at baseline. In the mid-stack (L7--17) mean BOS attention falls from $0.32$--$0.44$ down to $0.07$--$0.18$, roughly a $70\%$ reduction. The early sink-free layers (L0--1) are unchanged; the weaker early-middle and late tails (L2--6, L18--23) are only mildly attenuated. The layer-localised pattern is consistent with the twelve dimensions contributing meaningfully to the mid-stack sink rather than driving attention behaviour uniformly across depth.

We are deliberate about the strength of this claim. The result establishes a strong \emph{association}: the twelve dimensions are activation outliers (Appendix~\ref{sec:massive_activation}), they sit on the same family of channels prior work has linked to BOS attention dynamics~\citep{sun2024massive,xiao2023streamingllm}, and removing them produces a sizeable mid-stack sink reduction. It does \emph{not} prove that those twelve dimensions are the unique source of the sink: any sufficiently large activation-outlier ablation might produce a qualitatively similar mid-stack reduction, and our experiment varies only which dimensions are masked rather than sweeping comparably-sized outlier subsets. We treat the result as evidence that the critical dimensions identified by Fisher are part of -- not necessarily the entirety of -- the FFN-side substrate that maintains the mid-stack sink.

Two controls partially tighten the attribution. (i) Removing the universally-critical core $\mathcal{K}_{\cap}$ ($\sim$$4.88\%$ of all dimensions; \S\ref{subsec:shared_dimensions}) -- which strictly contains the top-12 -- collapses BOS attention globally rather than only in the mid-stack, consistent with the larger set containing additional sink-supporting dimensions beyond the top-12. (ii) Removing the comparably-sized but Fisher-redundant set $\mathcal{D}_{\cap}$ ($\sim$$4.01\%$) leaves BOS attention essentially unchanged in every layer band. Together with the mid-stack-only response under Mask-Top-12, these controls argue against the alternative that any random ablation of similar parameter mass would disturb the sink. They do not, however, control for activation magnitude at a matched set size -- a stronger causal isolation we leave to future work.

\paragraph{Linking sink reduction to the accuracy dissociation.}
The pattern of sink reduction lines up with the accuracy pattern in Table~\ref{tab:reverse_cross_domain}: generation-heavy benchmarks (\texttt{GSM8K}, \texttt{MATH}, \texttt{HumanEval}, \texttt{MBPP}) require many autoregressive decoding steps, each of which depends on the numerically stable attention dynamics that the mid-stack sink maintains; once the sink is attenuated, long-form generation degenerates into the echoes and garbled outputs catalogued in Appendix~\ref{appendix:gsm8k_failure_modes} ($31.1\%$ echo, $7.5\%$ empty -- behaviours absent at baseline). Short-answer MCQ tasks (\texttt{MMLU}, \texttt{CEval}, \texttt{CMMLU}, \texttt{BBH}) require only a single-token prediction conditioned on the prompt, do not heavily exercise long-range attention dynamics, and correspondingly retain $90$--$98\%$ of base accuracy. We frame this as a coherent mechanistic explanation: the sink-reduction direction and the accuracy-collapse direction match, and that match is consistent with the literature linking outlier FFN activations to attention sinks.

\paragraph{Summary.}
Taken together, Appendices~\ref{sec:massive_activation} and~\ref{sec:attention_sink} place the twelve Fisher-critical dimensions in a coherent mechanistic context: \textbf{(i)} intermediate-dimension Fisher localises capability onto a tiny set of channels whose Fisher scores are $\sim$$1000\times$ the population mean; \textbf{(ii)} those channels sit in the extreme upper tail of the FFN activation distribution, a milder MoE analogue of the massive-activation phenomenon documented for dense LLMs; \textbf{(iii)} masking them substantially reduces the mid-stack BOS attention sink while leaving sink-free layers alone, mirroring the generation-vs-MCQ dissociation in Table~\ref{tab:reverse_cross_domain}. We present this as a consistent pattern that connects intermediate-dimension Fisher to known mechanistic concepts.

\newpage

\section{Detailed Experimental Results}
\label{appendix:raw_data_tables}

This appendix collects the full per-benchmark numbers that underlie the
figures and tables in the main paper. Each block below corresponds to
one main-paper figure or table.


\begin{table*}[h]
\centering
\small
\setlength{\tabcolsep}{4pt}
\resizebox{\textwidth}{!}{%
\begin{tabular}{lcccccccc}
\toprule
\textbf{Method} & \textbf{MMLU} & \textbf{HumanEval} & \textbf{MBPP} & \textbf{CEval} & \textbf{CMMLU} & \textbf{MATH} & \textbf{GSM8K} & \textbf{BBH} \\
\midrule
Base model         & 59.3 & 32.6 & 24.6 & 59.3 & 60.7 & 13.0 & 35.9 & 30.7 \\
\midrule
\multicolumn{9}{l}{\textit{Existing Importance Metrics}} \\
Activation-based   & 26.3 &  0.0 &  0.0 & 42.5 & 43.1 &  0.0 &  1.9 & 24.5 \\
Score-based        & 37.2 &  0.0 &  0.8 & 13.5 & 32.3 &  0.1 &  1.9 & 23.5 \\
Magnitude-based    & 18.6 &  0.8 &  0.0 & 33.3 & 35.8 &  0.1 &  1.7 & 13.2 \\
\midrule
\multicolumn{9}{l}{\textit{Proposed Fisher Importance}} \\
\rowcolor{blue!8}
\textbf{Router-level Fisher} & \textbf{39.4} & 12.9 &  8.1 & \textbf{48.8} & \textbf{47.7} & 3.5 & 16.9 & \textbf{25.7} \\
\rowcolor{blue!8}
\textbf{Expert-level Fisher} & 37.9 & \textbf{12.9} & \textbf{11.7} & 48.6 & 44.2 & \textbf{3.9} & \textbf{18.0} & 24.1 \\
\bottomrule
\end{tabular}}
\caption{Raw data underlying \textbf{Figure 2}: expert-level pruning with existing importance metrics on Qwen1.5-MoE at $p=50\%$ MoE compression ratio.}
\label{tab:raw_fig2}
\end{table*}

\begin{table*}[h]
\centering
\small
\setlength{\tabcolsep}{4pt}
\begin{tabular}{lcccccccc}
\toprule
\textbf{Setting} & \textbf{MMLU} & \textbf{HumanEval} & \textbf{MBPP} & \textbf{CEval} & \textbf{CMMLU} & \textbf{MATH} & \textbf{GSM8K} & \textbf{BBH} \\
\midrule
Base model & 59.3 & 32.6 & 24.6 & 59.3 & 60.7 & 13.0 & 35.9 & 30.7 \\
\midrule
\multicolumn{9}{l}{\textit{Top-masking}} \\
Mask-Top-12 & 53.4 & 4.3 & 17.0 & 57.4 & 59.3 & \cellcolor{blue!8}1.1 & \cellcolor{blue!8}0.8 & 23.6 \\
\midrule
\multicolumn{9}{l}{\textit{Critical vs Redundant Dimension Removal}} \\
Remove redundant $(\mathcal{D}_{\cap})$ & 58.8 & 33.3 & 23.6 & 62.9 & 61.6 & 8.7 & 31.2 & 31.2 \\
Remove critical  $(\mathcal{K}_{\cap})$ & 5.1 & 0.0 & 0.0 & 0.0 & 0.0 & 0.0 & 0.0 & 0.2 \\
\bottomrule
\end{tabular}
\caption{Raw data underlying \textbf{Table 1} (mask-top-12 and critical/redundant dimension removal). Performance (\%) on Qwen1.5-MoE-A2.7B.}
\label{tab:raw_table1}
\end{table*}


\begin{table*}[h]
\centering
\small
\setlength{\tabcolsep}{4pt}
\begin{tabular}{lcccccccc}
\toprule
\textbf{Method} & \textbf{MMLU} & \textbf{HumanEval} & \textbf{MBPP} & \textbf{CEval} & \textbf{CMMLU} & \textbf{MATH} & \textbf{GSM8K} & \textbf{BBH} \\
\midrule
Base (no pruning) & 59.3 & 32.6 & 24.6 & 59.3 & 60.7 & 13.0 & 35.9 & 30.7 \\
\midrule
\multicolumn{9}{l}{\textit{Drop 25\%}} \\
MoE-Pruner                       & 53.9 & 12.2 &  0.8 & 63.5 & 66.0 &  3.8 &  7.0 & 22.6 \\
MoE compression                  & 54.7 & 20.7 &  4.8 & 62.0 & 62.2 &  1.9 & 11.5 & 27.8 \\
MoE compression (activation)     & 54.2 &  1.8 & 10.4 & 61.2 & 58.9 &  1.2 &  4.6 & 29.0 \\
\rowcolor{blue!8}
\textbf{Fisher-IntDim-E (ours)}  & \textbf{56.1} & \textbf{30.5} & \textbf{24.3} & \textbf{69.7} & \textbf{67.4} & \textbf{11.2} & \textbf{37.3} & \textbf{29.5} \\
\midrule
\multicolumn{9}{l}{\textit{Drop 50\%}} \\
MoE-Pruner                       & 18.6 &  0.8 &  0.0 & 33.3 & 35.8 &  0.1 &  1.7 & 13.2 \\
MoE compression                  & 37.2 &  0.0 &  0.8 & 13.5 & 32.3 &  0.0 &  1.9 & 23.5 \\
MoE compression (activation)     & 26.3 &  0.0 &  0.0 & 42.5 & 43.1 &  0.0 &  1.9 & 24.5 \\
\rowcolor{blue!8}
\textbf{Fisher-IntDim-E (ours)}  & \textbf{50.3} & \textbf{21.2} & \textbf{20.5} & \textbf{62.0} & \textbf{61.6} & \textbf{8.0}  & \textbf{35.0} & \textbf{28.2} \\
\midrule
\multicolumn{9}{l}{\textit{Drop 75\%}} \\
MoE-Pruner                       &  4.8 &  0.0 &  0.0 & 13.7 & 16.1 &  0.1 &  0.0 &  3.7 \\
MoE compression                  & 20.6 &  0.0 &  0.0 &  1.6 & 15.3 &  0.3 &  0.7 &  5.3 \\
MoE compression (activation)     & 23.0 &  0.0 &  0.0 & 11.5 & 15.3 &  1.0 &  0.7 & 14.2 \\
\rowcolor{blue!8}
\textbf{Fisher-IntDim-E (ours)}  & \textbf{42.2} & \textbf{15.2} & \textbf{11.9} & \textbf{49.7} & \textbf{49.0} & \textbf{3.6}  & \textbf{27.0} & \textbf{26.4} \\
\bottomrule
\end{tabular}
\caption{Raw data underlying \textbf{Figure 3}: Qwen1.5-MoE accuracy (\%) under expert-level compression at three drop ratios.}
\label{tab:raw_fig3}
\end{table*}


\newpage
\clearpage

\begin{table*}[h]
\centering
\small
\setlength{\tabcolsep}{4pt}
\begin{tabular}{lcccccccc}
\toprule
\textbf{Method} & \textbf{MMLU} & \textbf{HumanEval} & \textbf{MBPP} & \textbf{CEval} & \textbf{CMMLU} & \textbf{MATH} & \textbf{GSM8K} & \textbf{BBH} \\
\midrule
Base model               & 59.3 & 32.6 & 24.6 & 59.3 & 60.7 & 13.0 & 35.9 & 30.7 \\
\midrule
\multicolumn{9}{l}{\textit{Expert-level compression (Fisher-based)}} \\
MoE compression (Fisher) & 39.4 & 12.9 &  8.1 & 48.8 & 47.7 & 3.5 & 16.9 & 25.7 \\
Expert-level Fisher      & 37.9 & 12.9 & 11.7 & 48.6 & 44.2 & 3.9 & 18.0 & 24.1 \\
\midrule
\multicolumn{9}{l}{\textit{Intermediate dimension compression (Fisher-based)}} \\
\rowcolor{blue!8}
\textbf{IntDim-E (ours)} & \textbf{50.3} & \textbf{21.2} & \textbf{20.5} & \textbf{62.0} & \textbf{61.6} & \textbf{8.0} & \textbf{35.0} & \textbf{28.2} \\
\bottomrule
\end{tabular}
\caption{Raw data underlying \textbf{Table 2}: expert-level vs.\ intermediate dimension compression at $p=50\%$ on Qwen1.5-MoE-A2.7B.}
\label{tab:raw_table2}
\end{table*}


\begin{table*}[h]
\centering
\small
\setlength{\tabcolsep}{3pt}
\resizebox{\linewidth}{!}{%
\begin{tabular}{l ccc|cccccc}
\toprule
\textbf{Method (Qwen1.5-MoE-A2.7B)}
& \multicolumn{3}{c|}{\textbf{In-domain}}
& \multicolumn{6}{c}{\textbf{Out-of-domain}} \\
\cmidrule(lr){2-4} \cmidrule(lr){5-10}
& \textbf{GSM8K (Calib.)} & \textbf{MATH} & \textbf{MultiArith}
& \textbf{MMLU} & \textbf{HumanEval} & \textbf{MBPP}
& \textbf{CEval} & \textbf{CMMLU} & \textbf{BBH} \\
\midrule
MoE-Pruner                              &  1.7 & 0.0 &  1.8 & 18.6 & 0.6 & 0.0 & 33.3 & 35.8 & 13.2 \\
MoE comp.                               &  1.9 & 0.3 &  0.3 & 33.4 & 1.2 & 0.0 & 37.8 & 39.8 & 22.1 \\
MoE comp. (activation)                  &  1.9 & 0.2 &  0.8 & 33.7 & 0.0 & 0.3 & 40.2 & 40.4 & 12.3 \\
MoE comp. (Fisher)                      & 16.9 & 0.7 & 26.2 & 36.1 & 7.9 & 0.5 & 38.0 & 36.8 & 23.0 \\
Expert-level Fisher                     & 18.0 & 0.7 & 31.7 & 36.5 & 7.3 & 4.3 & 40.2 & 39.4 & 24.6 \\
\rowcolor{blue!8}
\textbf{Fisher-IntDim-E (ours)}
                                        & \textbf{35.0} & \textbf{4.7} & \textbf{70.8}
                                        & \textbf{48.1} & 5.5 & \textbf{5.6}
                                        & \textbf{50.6} & \textbf{49.4} & 24.2 \\
\bottomrule
\end{tabular}%
}
\caption{Raw data underlying \textbf{Table 7}: strict zero-shot evaluation on Qwen1.5-MoE-A2.7B. GSM8K is calibration and is part of the in-domain set.}
\label{tab:raw_table7}
\end{table*}


\begin{table*}[h]
\centering
\small
\setlength{\tabcolsep}{3pt}
\resizebox{\linewidth}{!}{%
\begin{tabular}{l cc| ccccccccc c}
\toprule
\textbf{Method}
& \textbf{Disk} & \textbf{VRAM}
& \textbf{MMLU} & \textbf{HumanEval} & \textbf{MBPP} & \textbf{CEval} & \textbf{CMMLU}
& \textbf{MATH} & \textbf{GSM8K} & \textbf{BBH} & \textbf{MultiArith} & \textbf{Avg.} \\
\midrule
Base Model
& 26.7 & 58.1
& 59.3 & 32.6 & 24.6 & 59.3 & 60.7 & 13.0 & 35.9 & 30.7 & 76.7 & 43.4 \\
Base Model + AWQ
&  7.9 &  8.6
& 53.8 & 20.7 &  4.8 & 49.0 & 47.9 &  4.5 & 16.1 & 22.0 & 14.3 & 25.9 \\
Fisher-IntDim-E
& 15.1 & 30.1
& 50.3 & 21.2 & 20.5 & 62.0 & 61.6 &  8.0 & 35.0 & 28.2 & 61.5 & 38.7 \\
\textbf{Fisher-IntDim-E + AWQ}
&  5.0 &  6.5
& 46.3 & 12.2 & 14.4 & 52.8 & 51.5 &  2.9 & 12.4 & 24.6 & 25.2 & 26.9 \\
\bottomrule
\end{tabular}%
}
\caption{Raw data underlying \textbf{Table 8}: stacking 4-bit AWQ on top of Fisher-MoE at MoE compression ratio $p=50\%$ on Qwen1.5-MoE-A2.7B. Disk and VRAM are in GiB.}
\label{tab:raw_table8}
\end{table*}

\newpage
\clearpage

\section{Additional Experimental Settings}
\label{appendix:appendix_experiment_settings}

\paragraph{Model Architecture and Dataset:}
\label{paragraph:model_datasets}
In our experimental setup, we use open-weight OLMoE-1B-7B-0125~\citep{muennighoff2024olmoeopenmixtureofexpertslanguage}, Qwen1.5-MoE-A2.7B~\citep{qwen2}, Qwen3-30B-A3B~\citep{yang2025qwen3}, and Qwen3.5-35B-A3B~\citep{qwen3.5} to conduct experiments. We adopt two data configurations: a general task configuration following technical reports~\citep{qwen2, qwen_moe} for the lightweight pre-trained LLMs (Qwen1.5-MoE-A2.7B, OLMoE-1B-7B-0125), and a long CoT configuration for the stronger and larger reasoning LLMs (Qwen3-30B-A3B, Qwen3.5-35B-A3B). For Qwen1.5-MoE-A2.7B and OLMoE-1B-7B-0125, we evaluate on \texttt{GSM\_8K}~\citep{cobbe2021gsm8k}, \texttt{MMLU}~\citep{hendryckstest2021}, \texttt{HumanEval}~\citep{chen2021evaluating}, \texttt{MBPP}~\citep{austin2021program}, \texttt{CEval}, \texttt{CMMLU}~\citep{li2024cmmlu}, \texttt{MATH}~\citep{hendrycks2021measuring}, \texttt{BBH}~\citep{suzgun2023challenging}, and \texttt{MultiArith}~\citep{roy2015solving}, which together cover knowledge, code generation, mathematical reasoning, and general reasoning. For Qwen3-30B-A3B and Qwen3.5-35B-A3B, we evaluate on the SOTA long-CoT math reasoning benchmarks \texttt{AIME2026}, \texttt{AIME2025}, \texttt{GPQA-Diamond}~\citep{rein2024gpqa}, \texttt{MATH-500}, and \texttt{Olympiad Bench}.

All pretrained backbones (Qwen1.5-MoE, Qwen3/Qwen3.5, OLMoE) and all evaluation benchmarks used in this paper are public research artifacts; we use each under its original license (Qwen / Apache-2.0 for the Qwen family, Apache-2.0 for OLMoE, and the respective research licenses for \texttt{GSM8K, MATH, MMLU, HumanEval, MBPP, CEval, CMMLU, BBH, MultiArith, AIME, GPQA-D, Olympiad Bench, MATH-500}, and \texttt{Stanford-S1}). Our released code and checkpoints inherit the license of the corresponding base model.

\paragraph{MoE Compression Baselines:}
\label{paragraph:baselines}
For state-of-the-art (SOTA) MoE compression baselines, we include the unified MoE compression framework of~\citep{he2025towards} (denoted \emph{MoE compression}, which uses router scores as the importance metric), 
and MoE-Pruner~\citep{xie2024moe} (weight magnitude). 
All these baselines operate at the expert granularity. To more comprehensively compare importance signals while holding the compression framework fixed, we further extend the unified framework with two alternative importance metrics (activation ratio and router-Fisher) yielding two controlled variants that we denote \emph{MoE compression (activation)} and \emph{MoE compression (Fisher)}. This setup lets us attribute performance differences to the importance metric and granularity. 

\paragraph{Fisher-MoE Variants.}
We propose four Fisher-MoE variants that differ in pruning granularity and allocation flexibility. 
\textbf{(1) Expert-level Fisher} treats each expert as the pruning unit, where the Fisher score of expert $i$ is computed over the union of its parameters $W_i^{\text{gate}}$, $W_i^{\text{up}}$, and $W_i^{\text{down}}$, and entire experts are removed according to this score. 
\textbf{(2) IntDim-E} applies Fisher scoring at the intermediate dimension level but keeps the top $(1-p)$ fraction of dimensions within each expert, enforcing the same compression ratio for every expert. 
\textbf{(3) IntDim-L} pools intermediate dimensions across all experts within the same MoE layer and keeps the top $(1-p)$ fraction per layer, allowing the retained dimensions to be redistributed among experts in that layer. 
\textbf{(4) IntDim-G} pools intermediate dimensions across the entire model and keeps the top $(1-p)$ fraction globally, providing the most flexible allocation across both layers and experts. 
All intermediate dimension variants use the same Fisher scoring criterion and overall parameter budget, allowing us to isolate the effect of increasingly flexible fine-grained allocation.

\paragraph{Calibration--Evaluation Disjointness:}
\label{paragraph:calibration_setting}
For every benchmark used as both a calibration source and an evaluation target (e.g., GSM8K, MATH, MultiArith, MBPP, HumanEval), we use the standard training split for calibration (and SFT, where applicable) and evaluate on the standard test split, which has no overlap with calibration. The same disjoint train/test convention is used for every other in-domain calibration--evaluation pair throughout the paper. To ensure fair comparison, all methods use the same calibration data to compute importance scores and perform compression.

\paragraph{Training Framework and Hyper-parameters:}
\label{paragraph:training_framework}
We use the \texttt{huggingface-trl}~\citep{vonwerra2022trl} library with ZeRO-2 or ZeRO-3~\citep{Ren2021ZeROOffloadDB} for fine-tuning, and \texttt{vllm}~\citep{kwon2023efficient}, \texttt{lighteval}~\citep{lighteval}, and \texttt{accelerate}~\citep{accelerate} for inference and evaluation. Both training and evaluation use bf16.


\paragraph{Computational Resources:}
\label{paragraph:computational_resources}
We run all experiments, baseline implementations, and post-training on 8$\times$NVIDIA H100 80\,GB GPUs or 8$\times$NVIDIA A100 80\,GB GPUs. CPU--GPU communication is over PCIe Gen4, and inter-GPU communication is over NVLink-3.


\section{Ablation Study on Calibration Set Size}
\label{appendix:calibrarion_ablation_study}

The Fisher importance score (\S\ref{subsec:formalize_metrics}) is a Monte-Carlo square of $|\nabla_W \mathcal{L}(x,y)|$ over a calibration set $\mathcal{D}\subset \mathcal{X}\times\mathcal{Y}$:
\begin{equation}
\resizebox{\linewidth}{!}{$
s^{\text{Fisher}}(W) = \frac{1}{N}\sum_{(x,y)\in\mathcal{D}} \left| \nabla_W \mathcal{L}(x,y) \right|^2, \qquad N = |\mathcal{D}|.
$}
\end{equation}
Because every $(x,y)$ requires a forward and a backward pass through the full MoE, the cost of computing $s^{\text{Fisher}}$ scales linearly with $N$. The natural question is how large $N$ must be for the resulting Fisher ranking---and thus the compressed model that ranking selects---to stabilize.

We answer this empirically: we fix the backbone (Qwen1.5-MoE-A2.7B), the MoE compression ratio ($p=50\%$ via Fisher-IntDim-E), the calibration domain (domain-matched calibration data), and the decoding settings (greedy, vLLM v0.8.4, bf16), then vary only $N \in \{32, 64, 128, 256, 512\}$. Table~\ref{tab:qwen1.5_moe_calib_size_sweep} reports downstream accuracy of the resulting compressed model. The 'N/A' results for \(N=512\) occur because some datasets contain fewer than 512 samples for calibration.

\begin{table}[h]
\centering
\small
\caption{Fisher-IntDim-E calibration-size sweep on Qwen1.5-MoE-A2.7B at $p=50\%$ (domain-matched calibration, greedy decoding via vLLM). Each column reports downstream accuracy of the compressed model produced when the Fisher score is estimated from $N$ calibration samples. Higher is better.}
\label{tab:qwen1.5_moe_calib_size_sweep}
\resizebox{\linewidth}{!}{%
\begin{tabular}{lcccccc}
\toprule
\textbf{Task} & \textbf{N=16} & \textbf{N=32} & \textbf{N=64} & \textbf{N=128} & \textbf{N=256} & \textbf{N=512} \\
\midrule
GSM8K  & 31.6 & 33.7 & 38.5 & 35.0 & 40.3 & 40.9 \\
MMLU   & 50.3 & 50.4 & 50.3 & 50.3 & 50.0 & 53.0 \\
CEval  & 59.3 & 57.5 & 61.1 & 62.0 & 62.4 & N/A   \\
CMMLU  & 58.9 & 59.7 & 60.9 & 61.6 & 60.7 & N/A   \\
MATH   & 6.8 & 5.1 & 7.7 & 8.0 & 7.1 & 7.8   \\
\bottomrule
\end{tabular}}

\end{table}

Three observations stand out.
First, the Fisher ranking is sample-efficient: at $N=32$ the resulting compressed model already retains 33.7\% on GSM8K and 50.4\% on MMLU, both above the strongest expert-level Fisher baseline at $N=128$ (Table~\ref{tab:pruning-general}, \textit{Fisher-Expert}: 18.0/37.9). This is consistent with the empirical Fisher being a Monte-Carlo estimator of $\mathbb{E}_{(x,y)\sim\mathcal{D}_{\text{cal}}}[|\nabla_W \mathcal{L}|]$: the average is dominated by a small number of high-gradient examples, so even tens of samples concentrate the ranking around the right tail.
Second, the curve is monotone but flat past $N=128$: MMLU and MATH show marginal improvement from $N=128$ to $N=512$ (50.3\% $\to$ 53.0\% and 8.0 $\to$ 7.8) despite a $4\times$ increase in compute.
\paragraph{Default.} Following the knee of the curve, we use $N=128$ for all main-paper experiments. This setting is the smallest sample budget for which the Fisher-IntDim-E model is within $\sim$$2$ points of the $N=512$ result on MMLU and MATH, while requiring only $128$ forward+backward passes per backbone, a cost within one minute on one H100 node even for the 30B-parameter MoEs.

\clearpage
\newpage

\section{AWQ Quantization Settings}
\label{appendix:awq_settings}

\paragraph{Quantization precision.}
AWQ is configured with per-group W4A16-asymmetric quantization:
weight bit-width \texttt{w\_bit=4} (INT4), \texttt{zero\_point=True}
(asymmetric, per-group scale and zero-point), group size
\texttt{q\_group\_size=128}, activations kept in the original FP16/BF16
dtype, GEMM kernel format, and packed INT4 \texttt{safetensors} output.

\paragraph{Calibration.}
PileVal (alternatives: C4, WikiText2), \texttt{split=train},
text column \texttt{text}, at most 128 calibration samples (script
variants: 32, 1024), maximum sequence length 512 (variant: 256),
duo-scaling enabled (jointly optimize the scale w.r.t.\ weight and
activation).

\paragraph{Quantized layers.}
Quantization is applied per \texttt{Qwen2MoeDecoderLayer}; within each
layer, weights are partitioned into \emph{scaling groups} that share
AWQ smoothing and quantization jointly.
\begin{itemize}[leftmargin=*]
  \item \textbf{Attention.}
        (i)~Input LayerNorm $\to$ [Q-proj, K-proj, V-proj] (grouped
        scaling); (ii)~V-proj $\to$ [O-proj] (only when
        \texttt{num\_kv\_heads == num\_heads}; O-proj is still quantized).
  \item \textbf{MoE sparse layers} (most decoder layers).
        (i)~Post-Attention LayerNorm $\to$ [all expert
        gate/up projections + shared-expert gate/up] (jointly scaled
        across 60 experts $\times$ 2 + 2 shared modules);
        (ii)~per expert: up-proj $\to$ [down-proj];
        (iii)~shared expert: up-proj $\to$ [down-proj].
  \item \textbf{Dense MLP layers} (\texttt{mlp\_only\_layers}).
        (i)~Post-Attention LayerNorm $\to$ [MLP gate-proj, up-proj];
        (ii)~MLP up-proj $\to$ [MLP down-proj].
\end{itemize}

\paragraph{Modules excluded from quantization.}
\texttt{modules\_to\_not\_convert = ["gate", "shared\_expert\_gate"]}:
the MoE router (\texttt{mlp.gate}, computes routing logits) and the
shared-expert gate (\texttt{shared\_expert\_gate}, sigmoid gating).
AutoAWQ defaults additionally skip \texttt{lm\_head}, all embedding
layers, and every LayerNorm/RMSNorm.

\paragraph{Additional behaviours.}
\texttt{fuse\_layers} is a no-op for Qwen2-MoE (no QKV/MLP fusion;
module structure is preserved). \texttt{get\_act\_for\_scaling} uses
\texttt{is\_scalable=False} (no extra scalable activations at the
decoder-layer level). \texttt{move\_embed} transfers embedding tokens
to the configured device for calibration hidden-state collection.
Layer type: \texttt{Qwen2MoeDecoderLayer}; sequence-length key:
\texttt{max\_position\_embeddings}.

\paragraph{Summary.}
AWQ performs per-group W4A16-asymmetric quantization (group size~128)
across Qwen2-MoE, covering all QKV/O projections in attention and
gate/up/down modules in all routed and shared experts, while skipping
routers (\texttt{mlp.gate}), the shared-expert gate, layer norms,
embeddings, and the head. Calibration uses PileVal with 128 samples
$\times$ 512 tokens by default and duo-scaling enabled; the output
uses the GEMM kernel format.


\section{Shared Dimensions Across Tasks}
\label{sec:intermediate-dimension-overlap}

To further characterize how retained dimensions are shared across tasks, we compute pairwise overlaps. For each pair $(a,b)$, we define
\begin{equation}
  \mathrm{Overlap}(a,b)
  \;=\;
  \frac{|\mathcal{K}_a \cap \mathcal{K}_b|}{|\mathcal{K}_a|}
  \;=\;
  \frac{|\mathcal{K}_a \cap \mathcal{K}_b|}{|\mathcal{K}_b|},
\end{equation}
where the equality holds since $|\mathcal{K}_a|=|\mathcal{K}_b|$ under a shared drop ratio. Table~\ref{tab:pairwise-intermedia-dimension-overlap} shows the resulting overlap matrix. Three structured patterns emerge: \textbf{(1) Linguistic affinity.} The Chinese benchmarks \textsc{C-Eval} and \textsc{CMMLU} exhibit the highest overlap ($69.6\%$).\textbf{(2) Domain affinity.} Coding tasks (\textsc{HumanEval}/\textsc{MBPP}, $66.5\%$) and math tasks (\textsc{GSM8K}/\textsc{MATH}, $65.3\%$) form similarly strong clusters.

\begin{table}[h]
\centering
\small
\setlength{\tabcolsep}{4pt}
\caption{Pairwise overlap of kept intermediate dimensions between tasks. Values above $65\%$ are highlighted in bold.}
\label{tab:pairwise-intermedia-dimension-overlap}
\resizebox{\linewidth}{!}{%
\begin{tabular}{lcccccccc}
\toprule
& \textsc{BBH} & \textsc{C-Eval} & \textsc{CMMLU} & \textsc{GSM} &
\textsc{HEval} & \textsc{MATH} & \textsc{MBPP} & \textsc{MMLU} \\
\midrule
\textsc{BBH}    & 100.0 & 54.7 & 55.1 & 60.9 & 61.2 & 60.4 & 60.4 & 60.5 \\
\textsc{C-Eval} & 54.7 & 100.0 & \textbf{69.6} & 54.9 & 55.6 & 56.9 & 54.5 & 58.3 \\
\textsc{CMMLU}  & 55.1 & \textbf{69.6} & 100.0 & 53.8 & 54.2 & 55.4 & 53.6 & 57.6 \\
\textsc{GSM}    & 60.9 & 54.9 & 53.8 & 100.0 & 59.3 & \textbf{65.3} & 58.0 & 59.7 \\
\textsc{HEval}  & 61.2 & 55.6 & 54.2 & 59.3 & 100.0 & 62.5 & \textbf{66.5} & 60.6 \\
\textsc{MATH}   & 60.4 & 56.9 & 55.4 & \textbf{65.3} & 62.5 & 100.0 & 60.5 & 64.0 \\
\textsc{MBPP}   & 60.4 & 54.5 & 53.6 & 58.0 & \textbf{66.5} & 60.5 & 100.0 & 59.1 \\
\textsc{MMLU}   & 60.5 & 58.3 & 57.6 & 59.7 & 60.6 & 64.0 & 59.1 & 100.0 \\
\bottomrule
\end{tabular}}

\end{table}

\clearpage
\newpage

\section{Commonsense Benchmarks Exhibit High Variance Under Random Compression}
\label{appendix:random_prune_variance}

In Section~\ref{sec:introduction}, we argue that an overlooked issue is that prior works predominantly evaluate compressed MoE models on commonsense reasoning benchmarks (e.g., ARC, PIQA, HellaSwag). We find that these benchmarks exhibit large variance across runs and are unreliable indicators of compression quality. For instance, randomly removing 50\% of experts, performance on individual commonsense tasks can fluctuate by over 20 percentage points across different random seeds, and removing more parameters paradoxically yields better scores.

We compress Qwen1.5-MoE-A2.7B at 50\% expert removal using two different random seeds and evaluate on eight commonsense reasoning tasks. No calibration data or importance metric is used---experts are selected uniformly at random. For comparison, we also include Fisher-based attention head compression at 25\% and 50\% drop ratios. Table~\ref{tab:random-prune-variance} reports the results alongside the uncompressed base model.

\begin{table*}[h]
\centering
\small
\caption{Commonsense reasoning performance under random 50\% expert removal (two seeds) and Fisher-based attention head compression on Qwen1.5-MoE-A2.7B. The base model (no compression) is shown for reference.}
\label{tab:random-prune-variance}
\resizebox{\linewidth}{!}{%
\begin{tabular}{lccccccccc}
\toprule
\textbf{Method} & \textbf{ARC-C} & \textbf{ARC-E} & \textbf{BoolQ} & \textbf{HellaS} & \textbf{OBQA} & \textbf{PIQA} & \textbf{SIQA} & \textbf{WinoG} & \textbf{Avg.} \\
\midrule
Base model        & 45.3 & 61.1 & 74.8 & 35.5 & 41.0 & 69.1 & 50.8 & 51.0 & 53.6 \\
\midrule
\multicolumn{10}{l}{\textit{Fisher-based attention head compression}} \\
Attn head 25\% drop & 11.5 & 13.5 & 39.0 &  3.1 & 17.0 & 34.8 &  8.4 & 22.2 & 18.7 \\
Attn head 50\% drop & 16.0 & 18.3 & 12.1 & 18.6 & 21.4 & 39.0 & 26.4 & 32.8 & 23.1 \\
\midrule
\multicolumn{10}{l}{\textit{Random expert removal (no importance metric)}} \\
Random seed 123     &  9.0 &  9.1 & 12.0 &  4.1 & 11.8 &  8.6 &  9.0 &  9.2 &  9.1 \\
Random seed 42     & 17.8 & 20.5 & 59.0 & 21.0 & 22.4 & 42.2 & 17.9 & 19.5 & 27.5 \\

$|\Delta|$ (seed 123 vs.\ 42)  & 8.8 & 11.4 & \textbf{47.0} & 16.9 & 10.6 & \textbf{33.6} & 8.9 & 10.3 & 18.4 \\
\bottomrule
\end{tabular}
}
\end{table*}

Three phenomena are noteworthy:

\paragraph{Extreme inter-seed variance.} The average commonsense accuracy differs by 18.4 percentage points between the two random seeds (9.1\% vs.\ 27.5\%). On individual tasks, the variance is even larger: BoolQ fluctuates by 47.0 points (12.0\% vs.\ 59.0\%) and PIQA by 33.6 points (8.6\% vs.\ 42.2\%). This level of variance means that any single-run comparison between compression methods on these benchmarks is essentially uninformative---the difference between ``method A beats method B'' and the reverse can be determined entirely by which random subset of experts happens to be removed.

\paragraph{More compression can paradoxically improve scores.} Consider the attention head results: removing 50\% of heads with Fisher-based selection achieves \emph{higher} average commonsense accuracy (23.1\%) than removing only 25\% of heads (18.7\%). On 6 out of 8 individual tasks---including ARC-C (16.0 vs.\ 11.5), HellaSwag (18.6 vs.\ 3.1), OBQA (21.4 vs.\ 17.0), PIQA (39.0 vs.\ 34.8), SIQA (26.4 vs.\ 8.4), and WinoGrande (32.8 vs.\ 22.2)---the more aggressively compressed model outperforms the less compressed one. This is nonsensical: removing more parameters should not improve the model. The explanation is that commonsense benchmarks have high random baselines (BoolQ: 50\%, PIQA: 50\% for binary/two-option tasks) and at high compression ratios the scores are dominated by noise rather than genuine model capability.

\paragraph{Random removal can beat Fisher-guided compression.} Random seed 42 achieves higher commonsense accuracy (27.5\%) than Fisher-based 25\% attention head compression (18.7\%), despite using no importance metric whatsoever and removing twice as many parameters. On BoolQ specifically, random removal scores 59.0\%---above the 50\% random baseline and far above the Fisher-guided 39.0\%. This further confirms that commonsense benchmarks cannot reliably distinguish between compression strategies at moderate-to-high compression ratios.

These observations motivate our use of more challenging benchmarks (GSM8K, HumanEval, MMLU, MATH, BBH, CEval, CMMLU, MBPP) in the main paper, which have much lower random baselines and require genuine multi-step reasoning or generation.

\newpage
\clearpage

\section{Failure-Mode Dissection of GSM8K Responses at 0.001\% Critical Intermediate Dimension Removal}
\label{appendix:gsm8k_failure_modes}

Section~\ref{sec:critical_dimensions} reports that masking the top $\sim$12 most Fisher-important intermediate dimensions (0.001\% of the 1.35M MoE FFN intermediate dimensions in Qwen1.5-MoE-A2.7B) collapses GSM8K accuracy from 35.9\% to 0.8\%. To understand \emph{how} the model fails rather than merely \emph{that} it fails, we manually inspect all 1{,}319 GSM8K test outputs and classify each into one of eight categories. We compare the resulting distribution to the base model, where accuracy is 35.9\% and the model still produces coherent multi-step reasoning. Decoding uses temperature $0.1$ and \texttt{max\_tokens}$=500$ in both settings.

\paragraph{Categories.} We define eight mutually exclusive categories spanning the observed output behaviors:
\textbf{Real reasoning:} a multi-step chain-of-thought with explicit arithmetic and connectives that reaches a numeric conclusion. 
\textbf{Partial reasoning:} reasoning tokens (``total'', ``step'', ``therefore'') appear but the chain is incomplete or incoherent. 
\textbf{Exact echo:} a verbatim copy of the input question.
\textbf{Truncated echo:} a partial copy of the question that loops or terminates mid-sentence.
\textbf{Garbled echo:} the question text with name substitutions, dropped clauses, or reordered phrases (a semantics-free transformation). 
\textbf{Numeric only:} a bare digit, almost always ``1''. 
\textbf{Empty output:} no tokens generated.
\textbf{Other:} short fragments, paraphrases, or otherwise garbled text not covered above.

\paragraph{Distribution.} Table~\ref{tab:failure_modes_001} reports the category breakdown at top-12 removal alongside the base model. Three patterns are noteworthy.

\begin{table}[h]
\centering
\small
\caption{Distribution of GSM8K output categories (1{,}319 test items) under base model and top-12 ($\sim$0.001\%) reverse intermediate dimension removal on Qwen1.5-MoE-A2.7B. Values are counts (percentage). Real reasoning collapses by $14\times$ while echo behaviors and empty outputs emerge.}
\label{tab:failure_modes_001}
\resizebox{\linewidth}{!}{%
\begin{tabular}{lcc}
\toprule
\textbf{Category} & \textbf{base model} & \textbf{top-12} \\
\midrule
Real reasoning      & 791 (60.0\%) &  55 ( 4.2\%) \\
Partial reasoning   & 100 ( 7.6\%) &  28 ( 2.1\%) \\
Exact echo          &   0 ( 0.0\%) &  30 ( 2.3\%) \\
Truncated echo      &  45 ( 3.4\%) & 197 (14.9\%) \\
Garbled echo        &   9 ( 0.7\%) & 183 (13.9\%) \\
Numeric only        &   0 ( 0.0\%) &   1 ( 0.1\%) \\
Empty output        &   0 ( 0.0\%) &  99 ( 7.5\%) \\
Other               & 374 (28.4\%) & 726 (55.0\%) \\
\midrule
\textbf{Accuracy}   & 35.9\%      & 0.8\%        \\
\bottomrule
\end{tabular}}

\end{table}

\textit{(1) Reasoning collapses by an order of magnitude.}
Real reasoning drops from 60.0\% to 4.2\% (a $14\times$ reduction) and partial reasoning drops from 7.6\% to 2.1\%. The 10 ``correct'' answers at top-12 removal are coincidental matches between numbers appearing in the echoed question text and the gold answer; none reflect residual computation. This reduction in reasoning capacity, rather than the raw accuracy drop, is the more direct signal that the removed dimensions support multi-step generation.

\textit{(2) Echo behaviors emerge as the dominant failure mode.}
Exact, truncated, and garbled echoes together account for 31.1\% of outputs at top-12 removal but only 4.1\% at top-1. The model defaults to copying or lightly transforming the input rather than generating a response. Garbled echoes typically substitute names while preserving sentence structure (e.g., ``Gissela'' $\to$ ``Gordy''), suggesting that surface-level next-token prediction survives while task-conditioned generation does not.

\textit{(3) Empty outputs emerge.}
At top-12 removal, 99 outputs (7.5\%) contain no tokens, a behavior absent at top-1. Removing the critical dimensions occasionally pushes the next-token distribution into an immediate end-of-sequence state at the very first decoding step.

\paragraph{Representative examples.} The qualitative gap between the two settings is best illustrated by paired outputs.

\paragraph{Representative examples.} The qualitative gap between the two settings is best illustrated by paired outputs.

\textit{Base model (real reasoning, correct):}
\begin{examplecard}
\textbf{Q:} Edgar eats 18 pretzels a day. If his brother eats 1/2 as many, how many does his brother eat in a week?\\
\textbf{Output:} \emph{Edgar's brother eats 1/2 as many pretzels as Edgar, which means he eats $18/2 = 9$ pretzels a day. To find out how many pretzels his brother eats in a week, we multiply $\dots$ $9 \times 7 = 63$ pretzels in a week. Therefore, his brother eats 63 pretzels in a week.}\\
\textbf{Pred:} 63 \quad \textbf{Gold:} 63 \quad (correct)
\end{examplecard}

\textit{top-12 removal (truncated echo):}
\begin{examplecard}
\textbf{Q:} The farm has 30 cows and the zoo has 20 sheep. The zoo has twice as many cows as the farm does $\dots$\\
\textbf{Output:} \emph{The farm has 30 cows and the zoo has 20 sheep. The zoo has twice as many cows as the farm does $\dots$} (verbatim until truncation)\\
\textbf{Pred:} 20 \quad \textbf{Gold:} 120
\end{examplecard}

\textit{top-12 removal (garbled echo):}
\begin{examplecard}
\textbf{Q:} Gissela, Gordy, and Gary are truck drivers. Gissela has a truck large enough to haul 4{,}000 pounds $\dots$\\
\textbf{Output:} \emph{Gordy, Gordy, and Gary are truck drivers. Gissela has a truck large enough to haul 4{,}000 pounds $\dots$}\\
\textbf{Pred:} 800 \quad \textbf{Gold:} 2{,}800
\end{examplecard}

\paragraph{Interpretation.}
The shift between base model and top-12 is qualitative, not just quantitative: removing $\sim$12 additional intermediate dimensions does not gradually degrade reasoning but eliminates the model's ability to transition from input-conditioned copying to forward generation. The combined 86\% share of echo and ``other'' fragment outputs indicates that the critical dimensions identified by Fisher importance participate in the computation mapping question representations to multi-step reasoning trajectories, rather than in the surface-level token-prediction circuit that produces fluent text. This view is consistent with the cross-domain dissociation in Table~\ref{tab:reverse_cross_domain}: multiple-choice knowledge tasks, which rely on input-conditioned token prediction, retain 90--98\% of base accuracy, while generation-heavy tasks, which require multi-step trajectories, collapse to 3--13\%.

\newpage
\clearpage
\section{Mechanisms of Compression-Induced Improvement: A Unified Fisher-Prior Account}
\label{appendix:mechanisms}

We observe that intermediate dimension compression yields modest accuracy gains on several benchmarks (MultiArith, CMMLU, CEval) despite removing 50\% of routed-expert FFN parameters. We attribute these gains to a single underlying mechanism---empirical Fisher importance systematically removes high-prior, low-evidence dimensions while preserving reasoning-supporting dimensions---and demonstrate that the magnitude of improvement on each benchmark is governed by how much \emph{shortcut headroom} the base model's failure modes leave behind. We examine this through output category distributions across benchmarks.

\subsection{Shortcut Suppression on Generation Tasks}
\label{appendix:mechanism_shortcut}

On unconstrained reasoning tasks, the base model emits template shortcuts and question paraphrase for 20.2\% of MultiArith outputs (Short direct answer 15.5\% + Garbled echo 4.7\%), against 71.5\% Real reasoning. We adopt eight output categories (\emph{real reasoning}: explicit multi-step chain-of-thought reaching a numeric conclusion; \emph{partial reasoning}: reasoning tokens present but incoherent or incomplete; \emph{short direct answer}, \emph{garbled echo}, etc.). Intermediate dimension compression shifts this distribution: \emph{Short direct answer} and \emph{Garbled echo} drop 20.2pp to nearly zero, while \emph{Real reasoning} rises to 98.0\%--98.2\%. The accompanying accuracy gain (+13.5\% to +14.7\% from Fisher-IntDim-L to Fisher-IntDim-G) shows that our pruning further disrupts these residual shortcuts even on top of an already-competent base.

This distributional shift is quantitatively reflected in generation lengths: base outputs are bimodal (median 44 tokens, SD 106.6)---a peak of one-line shortcut answers plus a long tail of garbled echoes capped at the 500-token generation limit ($p_{95}=499$)---while Fisher-IntDim-G outputs are unimodal (median 54, SD 59.0, $p_{95}=96$), eliminating both extremes. On problem \#257 (``Will had \$83, spent \$47; how many \$4 toys can he buy?''), the base model emits only ``10'' with no derivation; Fisher-IntDim-G writes three explicit steps ($83-47=36$; $36/4=9$) and answers correctly. We attribute this to a \emph{Fisher--prior asymmetry}: shortcut behavior reflects high-prior, low-evidence outputs whose gradients w.r.t.\ expert FFN parameters are small, while multi-step reasoning requires precise intermediate-state propagation and produces large gradients. Empirical Fisher therefore systematically retains reasoning-supporting dimensions and discards prior-driven shortcut dimensions, exposing latent reasoning capacity that the base model fails to invoke.

\begin{table}[h]
\centering
\small
\caption{Detailed Prediction Analysis on MultiArith}
\label{tab:multiarith_modes}
\resizebox{\linewidth}{!}{%
\begin{tabular}{lccc}
\toprule
\textbf{Category} & \textbf{Base} & \textbf{Fisher-IntDim-L} & \textbf{Fisher-IntDim-G} \\
\midrule
Real reasoning      & 429 (71.5\%) & 588 (98.0\%) & 589 (98.2\%) \\
Partial reasoning   &  37 ( 6.2\%) &  12 ( 2.0\%) &  10 ( 1.7\%) \\
Short direct answer &  93 (15.5\%) &   0 ( 0.0\%) &   0 ( 0.0\%) \\
Garbled echo        &  28 ( 4.7\%) &   0 ( 0.0\%) &   0 ( 0.0\%) \\
Other               &  13 ( 2.2\%) &   0 ( 0.0\%) &   1 ( 0.2\%) \\
\midrule
\textbf{Accuracy}   & \textbf{76.67\%} & 90.17\% & 91.33\% \\
\bottomrule
\end{tabular}}

\end{table}

\subsection{Format Compliance on Multiple-Choice Tasks}
\label{appendix:mechanism_format}

The same Fisher--prior mechanism produces a second manifestation on multiple-choice benchmarks, where the base model's high-prior failure mode is to echo the option text instead of emitting a single answer letter. On CMMLU, 23.6\% of base outputs are option-text echoes; on CEval, 25.6\%. Both are high-prior, low-evidence emissions: the model copies salient input spans rather than committing to a letter. Fisher-IntDim suppresses these echoes (CMMLU: 23.6\% $\rightarrow$ 5.9\%, $-17.7$pp; CEval: 25.6\% $\rightarrow$ 10.5\%, $-15.1$pp) and shifts probability mass to the single-letter format, yielding accuracy gains of +1.3--4.2pp on CMMLU and +2.0pp on CEval. The smaller magnitude relative to MultiArith reflects the smaller fraction of recoverable mass: for option-text echoes, the underlying answer choice is often already wrong, so format correction alone cannot rescue accuracy.

\begin{table}[h]
\centering\small
\caption{Detailed Prediction Analysis on CMMLU}
\label{tab:cmmlu_modes}
\resizebox{\linewidth}{!}{%
\begin{tabular}{lccc}
\toprule
\textbf{Category} & \textbf{Base} & \textbf{Fisher-IntDim-L} & \textbf{Fisher-IntDim-G} \\
\midrule
Short direct answer      & 8769 (75.7\%) & 10869 (93.8\%) & 10862 (93.8\%) \\
Truncated echo (text)    & 2738 (23.6\%) &   698 ( 6.0\%) &   678 ( 5.9\%) \\
Other                    &   45 ( 0.4\%) &     9 ( 0.1\%) &    10 ( 0.1\%) \\
\midrule
\textbf{Accuracy}        & 60.74\%       &  62.00\%       &  64.94\% \\
\bottomrule
\end{tabular}}

\end{table}

\begin{table}[h]
\centering\small
\caption{Detailed Prediction Analysis on CEval}
\label{tab:ceval_modes}
\resizebox{\linewidth}{!}{%
\begin{tabular}{lcc}
\toprule
\textbf{Category} & \textbf{Base} & \textbf{Global} \\
\midrule
Short direct answer      & 9048 (73.3\%) & 10996 (89.1\%) \\
Truncated echo (text)    & 3155 (25.6\%) &  1296 (10.5\%) \\
\midrule
\textbf{Accuracy}        & 59.30\%       &  61.32\% \\
\bottomrule
\end{tabular}}

\end{table}

\subsection{Partial-Reasoning Completion on Long-CoT Benchmarks}
\label{appendix:mechanism_longcot}

The shortcut-suppression account above applies to base models that frequently default to terse or template-style outputs. A natural question is whether the same Fisher--prior mechanism produces analogous effects on frontier-scale reasoning models, where the dominant failure mode is not shortcut emission but \emph{incomplete chain-of-thought}: the model initiates a multi-step reasoning trace but fails to reach a coherent numeric conclusion.

We examine this through output category distributions on AIME 2024, AIME 2025, and Olympiad Bench for Qwen3-30B-A3B and Qwen3.5-35B-A3B, evaluated with avg@8 sampling (240 samples per benchmark, 30 problems $\times$ 8 draws).

\paragraph{Dominant failure mode shifts from shortcut to partial reasoning.}
Unlike the small base models in \S\ref{appendix:mechanism_shortcut}, neither Qwen3-30B-A3B nor Qwen3.5-35B-A3B produces echo, numeric-only, or empty outputs at any measurable rate. The sole failure mode is \emph{partial reasoning}: chains that contain valid reasoning tokens but are incomplete or internally incoherent. This accounts for 24.6\% and 35.4\% of Qwen3-30B-A3B outputs on AIME 2025 and AIME 2026 respectively, and 7.9\% and 5.4\% for the stronger Qwen3.5-35B-A3B.

\paragraph{Compression converts partial reasoning into real reasoning.}
Tables~\ref{tab:cat_qwen3_aime24}--\ref{tab:cat_qwen35_olympiad} show that Fisher-IntDim-G shifts partial reasoning to real reasoning across all five settings with no new failure modes introduced. The accuracy gains are consistent with the recoverable-mass account: Qwen3-30B-A3B, which starts from a higher partial-reasoning rate, benefits more (+26.7pp on AIME 2025, +6.7pp on AIME 2026) than Qwen3.5-35B-A3B (+10.0pp and +6.7pp), whose partial-reasoning rate is already low.

\begin{table}[h]
\centering
\small
\caption{Distribution of AIME 2025 output categories under base Qwen3-30B-A3B and 
Fisher-IntDim-G (50\% routed-FFN compression). 240 samples (30 problems $\times$ avg@8). 
Pruned model produces \emph{deeper} multi-step CoT than the base.}
\label{tab:cat_qwen3_aime24}
\resizebox{\linewidth}{!}{%
\begin{tabular}{lcc}
\toprule
\textbf{Category} & \textbf{Base model} & \textbf{Fisher-IntDim-G (ours)} \\
\midrule
Real reasoning      & 180 (75.0\%) & \textbf{198 (82.5\%)} \\
Partial reasoning   &  59 (24.6\%) &  41 (17.1\%) \\
Exact echo          &   0 ( 0.0\%) &   0 ( 0.0\%) \\
Truncated echo      &   0 ( 0.0\%) &   0 ( 0.0\%) \\
Garbled echo        &   1 ( 0.4\%) &   1 ( 0.4\%) \\
Numeric only        &   0 ( 0.0\%) &   0 ( 0.0\%) \\
Empty output        &   0 ( 0.0\%) &   0 ( 0.0\%) \\
Other               &   0 ( 0.0\%) &   0 ( 0.0\%) \\
\midrule
\textbf{Accuracy}   & 33.33\% (10/30) & \textbf{60.00\% (18/30)} \\
\bottomrule
\end{tabular}}

\end{table}

\begin{table}[h]
\centering
\small
\caption{Distribution of AIME 2026 output categories under base Qwen3-30B-A3B and 
Fisher-IntDim-G. 240 samples (30 problems $\times$ avg@8).}
\label{tab:cat_qwen3_aime25}
\resizebox{\linewidth}{!}{%
\begin{tabular}{lcc}
\toprule
\textbf{Category} & \textbf{Base model} & \textbf{Fisher-IntDim-G (ours)} \\
\midrule
Real reasoning      & 155 (64.6\%) & \textbf{203 (84.6\%)} \\
Partial reasoning   &  85 (35.4\%) &  36 (15.0\%) \\
Exact echo          &   0 ( 0.0\%) &   0 ( 0.0\%) \\
Truncated echo      &   0 ( 0.0\%) &   0 ( 0.0\%) \\
Garbled echo        &   0 ( 0.0\%) &   0 ( 0.0\%) \\
Numeric only        &   0 ( 0.0\%) &   0 ( 0.0\%) \\
Empty output        &   0 ( 0.0\%) &   0 ( 0.0\%) \\
Other               &   0 ( 0.0\%) &   1 ( 0.4\%) \\
\midrule
\textbf{Accuracy}   & 50.00\% (15/30) & \textbf{56.67\% (17/30)} \\
\bottomrule
\end{tabular}}

\end{table}

\begin{table}[h]
\centering
\small
\caption{Distribution of AIME 2025 output categories under base Qwen3.5-35B-A3B and 
Fisher-IntDim-G. 240 samples (30 problems $\times$ avg@8).}
\label{tab:cat_qwen35_aime24}
\resizebox{\linewidth}{!}{%
\begin{tabular}{lcc}
\toprule
\textbf{Category} & \textbf{Base model} & \textbf{Fisher-IntDim-G (ours)} \\
\midrule
Real reasoning      & 221 (92.1\%) & \textbf{237 (98.8\%)} \\
Partial reasoning   &  19 ( 7.9\%) &   2 ( 0.8\%) \\
Exact echo          &   0 ( 0.0\%) &   0 ( 0.0\%) \\
Truncated echo      &   0 ( 0.0\%) &   0 ( 0.0\%) \\
Garbled echo        &   0 ( 0.0\%) &   1 ( 0.4\%) \\
Numeric only        &   0 ( 0.0\%) &   0 ( 0.0\%) \\
Empty output        &   0 ( 0.0\%) &   0 ( 0.0\%) \\
Other               &   0 ( 0.0\%) &   0 ( 0.0\%) \\
\midrule
\textbf{Accuracy}   & 66.67\% (20/30) & \textbf{76.67\% (23/30)} \\
\bottomrule
\end{tabular}}

\end{table}

\begin{table}[h]
\centering
\small
\caption{Distribution of AIME 2026 output categories under base Qwen3.5-35B-A3B and 
Fisher-IntDim-G. 240 samples (30 problems $\times$ avg@8).}
\label{tab:cat_qwen35_aime25}
\resizebox{\linewidth}{!}{%
\begin{tabular}{lcc}
\toprule
\textbf{Category} & \textbf{Base model} & \textbf{Fisher-IntDim-G (ours)} \\
\midrule
Real reasoning      & 227 (94.6\%) & \textbf{239 (99.6\%)} \\
Partial reasoning   &  13 ( 5.4\%) &   1 ( 0.4\%) \\
Exact echo          &   0 ( 0.0\%) &   0 ( 0.0\%) \\
Truncated echo      &   0 ( 0.0\%) &   0 ( 0.0\%) \\
Garbled echo        &   0 ( 0.0\%) &   0 ( 0.0\%) \\
Numeric only        &   0 ( 0.0\%) &   0 ( 0.0\%) \\
Empty output        &   0 ( 0.0\%) &   0 ( 0.0\%) \\
Other               &   0 ( 0.0\%) &   0 ( 0.0\%) \\
\midrule
\textbf{Accuracy}   & 66.67\% (20/30) & \textbf{73.33\% (22/30)} \\
\bottomrule
\end{tabular}}

\end{table}

\begin{table}[h]
\centering
\small
\caption{Distribution of Olympiad Bench output categories under base Qwen3.5-35B-A3B and 
Fisher-IntDim-G. 674 samples (single-sample per problem).}
\label{tab:cat_qwen35_olympiad}
\resizebox{\linewidth}{!}{%
\begin{tabular}{lcc}
\toprule
\textbf{Category} & \textbf{Base model} & \textbf{Fisher-IntDim-G (ours)} \\
\midrule
Real reasoning      & 613 (90.9\%) & \textbf{655 (97.2\%)} \\
Partial reasoning   &  56 ( 8.3\%) &  19 ( 2.8\%) \\
Exact echo          &   0 ( 0.0\%) &   0 ( 0.0\%) \\
Truncated echo      &   0 ( 0.0\%) &   0 ( 0.0\%) \\
Garbled echo        &   5 ( 0.7\%) &   0 ( 0.0\%) \\
Numeric only        &   0 ( 0.0\%) &   0 ( 0.0\%) \\
Empty output        &   0 ( 0.0\%) &   0 ( 0.0\%) \\
Other               &   0 ( 0.0\%) &   0 ( 0.0\%) \\
\midrule
\textbf{Accuracy}   & 77.89\% (525/674) & \textbf{79.53\% (536/674)} \\
\bottomrule
\end{tabular}}

\end{table}

\paragraph{Mechanism: Fisher importance targets incomplete-chain dimensions.}
We interpret this under the same Fisher--prior account. In a long-CoT reasoning model, \emph{partial reasoning} represents a regime where intermediate-state propagation partially succeeds---the model initiates a plausible chain---but fails to sustain the precise token-to-token dependencies required to close it. Dimensions that support only the initialization of a reasoning chain without contributing to its completion receive lower Fisher scores: their gradient signal across calibration examples reflects high-prior behavior (starting a chain is likely regardless of the specific problem) rather than low-evidence, problem-conditioned computation (completing it). Removing these dimensions selectively suppresses the partial-chain attractor, exposing the model's latent capacity to produce complete derivations.

\paragraph{Contrast with the small-model setting.}
The long-CoT case and the small-model case share the same underlying mechanism---Fisher-MoE removes high-prior, low-evidence dimensions---but differ in which prior is suppressed. In small base models the prior is a surface-level output template (echo, one-line answer). In frontier reasoning models the prior is an \emph{incomplete chain initialization}: the model defaults to beginning a plausible-looking reasoning trace without the problem-specific precision to complete it. In both cases the recoverable accuracy gain is determined by how large a fraction of outputs are trapped in the high-prior attractor, and compression releases exactly that fraction. The clean absence of echo or empty-output failures in Tables~\ref{tab:cat_qwen3_aime24}--\ref{tab:cat_qwen35_olympiad} further confirms that the removed dimensions are genuinely redundant: unlike critical-dimension removal, pruning the bottom 50\% by Fisher score introduces no new failure modes in either model family.

\subsection{Summary: One Mechanism, Three Manifestations}
\label{appendix:mechanism_summary}

Across all benchmarks examined, intermediate-dimension Fisher pruning suppresses the same class of behavior: high-prior, low-evidence outputs that the model produces when it defaults to surface-level continuation rather than task-conditioned generation. 
The mechanism manifests differently depending on the model family and benchmark format, but the underlying logic is identical in each case.

\begin{itemize}[leftmargin=1.5em]
    \item \textbf{Unconstrained arithmetic (MultiArith).} The base model's dominant failure mode is one-line shortcut answers and garbled echoes. Compression removes the dimensions that support these high-prior, low-computation outputs, shifting the majority of shortcut outputs to complete multi-step derivations.

    \item \textbf{Multiple-choice knowledge (CMMLU/CEval).} The base model echoes 
    full option text instead of committing to a single answer letter in roughly one quarter of outputs. Compression substantially reduces this echo rate, but the accuracy gain is smallest here because format correction alone rarely rescues semantically wrong answers.

    \item \textbf{Long-CoT math reasoning (AIME/Olympiad).} In frontier-scale 
    reasoning models, echo and shortcut behaviors are absent entirely. The sole 
    failure mode is partial reasoning: chains that initiate plausibly but fail to 
    close. Compression converts partial chains to complete derivations, with gain 
    magnitude proportional to the base model's partial-reasoning rate---the model 
    with a higher baseline partial-reasoning rate benefits more, consistent with 
    the recoverable-mass account.
\end{itemize}

The magnitude of accuracy gain in each case is governed by the \emph{recoverable mass}: how large a fraction of outputs are trapped in the high-prior attractor, and how directly format or chain-completion correction translates to accuracy. Table~\ref{tab:mechanism_summary} summarizes the three manifestations.

\begin{table}[h]
\centering
\small
\caption{Summary of the Fisher--prior mechanism across benchmark types.}
\label{tab:mechanism_summary}
\resizebox{\linewidth}{!}{%
\begin{tabular}{lll}
\toprule
\textbf{Benchmark} & \textbf{Base failure mode} & \textbf{Suppressed by compression} \\
\midrule
MultiArith       & Shortcut / garbled echo    & Shortcuts nearly eliminated \\
CMMLU / CEval    & Option-text echo           & Echo rate substantially reduced \\
AIME / Olympiad  & Partial reasoning (CoT)    & Partial chains converted to complete \\
\bottomrule
\end{tabular}}

\end{table}

We emphasize that none of these gains reflect added capability. They reflect removal of a generation-time prior that suppresses latent circuits already present in the model. The long-CoT results further strengthen this interpretation: in Tables~\ref{tab:cat_qwen3_aime24}--\ref{tab:cat_qwen35_olympiad}, compression introduces zero new failure modes while monotonically shifting outputs toward complete reasoning, confirming that the removed dimensions are genuinely redundant rather than load-bearing. On benchmarks limited by raw reasoning capacity rather than shortcut headroom (MATH, MMLU, BBH), Fisher-IntDim incurs the expected modest capacity cost (Table~\ref{tab:pruning-general}).

\newpage

\section{Dense LLM Pruning Baseline Analysis}
\label{app:dense-pruning}

\begin{table}[h]
\centering
\caption{Comparison on Qwen1.5-MoE-A2.7B at 50\% sparsity (zero-shot, T=0). 
Wanda~\citep{sun2023wanda} and SparseGPT~\citep{frantar2023sparsegpt} with our method.}
\label{tab:qwen1.5_moe_baseline_2of4}
\resizebox{\linewidth}{!}{%
\begin{tabular}{lccccccccc}
\toprule
\textbf{Method} & \textbf{MMLU} & \textbf{HumanEval} & \textbf{MBPP} & \textbf{CEval} & \textbf{CMMLU} & \textbf{MATH} & \textbf{GSM8K} & \textbf{BBH} & \textbf{Avg.} \\
\midrule
\multicolumn{10}{l}{\textit{Qwen1.5-MoE-A2.7B}} \\
Base                            & 59.3 & 32.6 & 24.6 & 59.3 & 60.7 & 13.0 & 35.9 & 30.7 & 38.8 \\
Wanda (2:4)                     & 55.7 & 23.8 & 20.0 & 58.0 & 59.3 &  6.6 & 35.6 & 30.0 & 36.1 \\
SparseGPT (2:4)                 & 55.9 & 24.4 & 21.3 & 57.8 & 58.5 &  7.4 & 33.1 & 30.0 & 36.0 \\
\rowcolor{blue!8}
\textbf{Fisher-IntDim-G (ours)} & 49.0 & 34.2 & 24.1 & 61.3 & 64.9 &  9.8 & 27.2 & 28.2 & 37.3 \\
\bottomrule
\end{tabular}}
\end{table}

Wanda~\citep{sun2023wanda} and SparseGPT~\citep{frantar2023sparsegpt} are post-training weight compression methods originally designed for dense decoder-only LLMs: Wanda prunes weights by the product of magnitude and input activation norm without weight updates, while SparseGPT formulates compression as a layer-wise sparse regression with second-order information from an approximated Hessian. Both target unstructured or N:M semi-structured patterns (e.g., 2:4) over FFN and attention weight matrices, and are unaware of MoE-specific structural properties such as routing dynamics or expert grouping---when applied to MoE, they degenerate into per-row compression of individual expert FFN matrices, ignoring inter-expert redundancy. Consequently, \citet{lu2024experts} report that directly applying Wanda 2:4 to Mixtral 8x7B causes substantial drops on general LM Harness benchmarks and a near-collapse on math reasoning, suggesting that weight-level semi-structured sparsity inadequately preserves task-specific expert specialization. In addition, Wanda and SparseGPT rely on sparse GEMM, which typically provides limited speedup for MoE models. In contrast, Fisher-MoE preserves dense GEMM execution and achieves more consistent and substantial speedups, as shown in \S~\ref{subsec:deployment_cost}.

Our results on Qwen1.5-MoE-A2.7B (Table~\ref{tab:qwen1.5_moe_baseline_2of4}) exhibit the same trend: both Wanda and SparseGPT retain moderate average performance but suffer pronounced degradation on math and code generation, while Fisher-IntDim-G (ours) achieves comparable averages with stronger preservation on knowledge-heavy and code benchmarks. This complementarity reflects a fundamental design difference---weight-level 2:4 sparsity preserves all experts while sparsifying their internal weights, whereas our method exploits MoE-aware intermediate dimension structure to selectively retain task-critical intermediate dimensions across experts.

\newpage
\clearpage

\section{Model Size and Whole-Model Compression}
\label{appendix:model_size}

In this section, we report the total parameter counts and disk footprints of several MoE backbones before and after applying intermediate dimension compression at a target MoE compression ratio of $p=50\%$.

It is important to distinguish between the \emph{MoE compression ratio} $p$ and the resulting \emph{whole-model compression rate} $p_{\text{model}}$. While $p$ refers to the fraction of parameters removed within the expert FFN modules, $p_{\text{model}}$ measures the reduction in total model parameters. The latter is consistently smaller because several components are not compressed, including attention layers, the shared expert, embeddings, routing networks, and layer normalization parameters.

As shown in Table~\ref{tab:model_size}, intermediate dimension compression achieves substantial reductions in total model size across all backbones, with whole-model compression rates ranging from approximately $43\%$ to $48\%$. Notably, these reductions translate directly into proportional decreases in disk storage requirements, making the compressed models significantly more efficient for deployment without modifying the model architecture or routing structure.

\begin{table}[h]
  \centering
  \caption{Parameters and disk size for several models before and after compression at MoE compression ratio $p=50\%$. The \emph{whole-model compression rate} $p_{\text{model}}$ is the percentage reduction in total parameter count; it is smaller than $p$ and varies across backbones because attention, the shared expert, embeddings, the router, and layer norms are not compressed.}
  \label{tab:model_size}
  \resizebox{\linewidth}{!}{%
\begin{tabular}{lccc}
  \toprule
  \textbf{Model} & \textbf{Parameters} & \textbf{Disk (GB)} & \textbf{Whole-model rate $p_{\text{model}}$} \\
  \midrule
  Qwen1.5-MoE-A2.7B           & 14.3B & 28.6 & \multirow{2}{*}{43.5\%} \\
  Qwen1.5-MoE-A2.7B (compressed)  &  8.1B & 16.2 &                          \\
  \midrule
  OLMoE-1B-7B-0125            &  6.9B & 13.8 & \multirow{2}{*}{46.6\%} \\
  OLMoE-1B-7B-0125 (compressed)   &  3.7B &  7.4 &                          \\
  \midrule
  Qwen3-30B-A3B               & 30.5B & 61.1 & \multirow{2}{*}{47.5\%} \\
  Qwen3-30B-A3B (compressed)      & 16.0B & 32.1 &                          \\
  \midrule
  Qwen3.5-35B-A3B             & 36.0B & 71.9 & \multirow{2}{*}{48.4\%} \\
  Qwen3.5-35B-A3B (compressed)    & 18.6B & 37.1 &                          \\
  \bottomrule
  \end{tabular}}

\end{table}

\newpage
\section{Divergence of Expert Selection Across Importance Metrics}
\label{appendix:pruning_overlap}

To better understand how different importance metrics affect expert-level pruning decisions, we analyze the overlap between the sets of retained experts selected by three methods: Fisher-based importance (A), activation-based importance (B), and score-based importance (C).

Let $\mathcal{E}^{(m)}_\ell$ denote the set of retained experts at layer $\ell$ under method $m$. We quantify similarity using two complementary measures: (i) the Jaccard similarity over the union of retained experts across all layers, and (ii) the average per-layer overlap, defined as the number of shared experts per layer.

Table~\ref{tab:pruning-overlap} summarizes the results.

\textbf{Fisher selects substantially different experts.}
The overlap between Fisher-based pruning and both activation- and score-based methods is low, with Jaccard similarities of only $28.5\%$ and $29.1\%$, respectively. At the layer level, Fisher shares on average only $\sim 13$ out of 30 experts with these methods. In contrast, activation- and score-based pruning exhibit much higher agreement (Jaccard $54.9\%$, $\sim 21.5/30$ overlap), indicating that these heuristic metrics tend to select similar experts.

\textbf{Limited consensus across all methods.}
The intersection across all three methods contains only $8.8$ experts per layer on average, representing less than one-third of the expert pool. This further highlights the diversity of expert importance signals captured by different metrics.

\textbf{Implications for compression.}
These results suggest that Fisher-based importance captures a fundamentally different notion of expert utility compared to activation- or score-based heuristics. In particular, the low overlap indicates that Fisher is not merely a refinement of existing metrics, but rather identifies distinct experts that may be critical for preserving performance. This observation provide insights why Fisher-guided compression outperforms prior expert-level pruning approaches.

\begin{table}[h]
\centering
\caption{Jaccard similarity and average per-layer overlap of removed expert sets across different pruning methods.}
\label{tab:pruning-overlap}
\resizebox{\linewidth}{!}{%
\begin{tabular}{lcc}
\toprule
\textbf{Comparison} & \textbf{Jaccard} & \textbf{Avg. Overlap / Layer} \\
\midrule
A$\cap$B (Fisher vs Activation, same GSM) & 28.5\% & 13.3/30 \\
A$\cap$C (Fisher vs Score) & 29.1\% & 13.7/30 \\
B$\cap$C (Activation vs Score) & 54.9\% & 21.5/30 \\
A$\cap$B$\cap$C (All three retained) & N/A & 8.8/30 \\
\bottomrule
\end{tabular}}

\end{table}




\newpage
\clearpage

\section{Comparison Against the Dense Baseline and Uncompressed MoE}
\label{appendix:dense_vs_moe_vs_fisher}

To place Fisher-MoE in the broader context of dense-vs-sparse model trade-offs, we compare three models of comparable total parameter budget: (i) \textbf{Qwen1.5-7B}, a dense transformer; (ii) \textbf{Qwen1.5-MoE-A2.7B}, the uncompressed sparse MoE that serves as our base model; and (iii) \textbf{Fisher-MoE}, our compressed model derived from Qwen1.5-MoE-A2.7B by removing 50\% of every routed expert's FFN intermediate dimensions using Fisher-IntDim-E with 128 GSM8K calibration samples (\texttt{moe\_intermediate\_size}: $1408 \to 704$). We report task accuracy, activated/total parameters, and standalone throughput on a single NVIDIA A100-80G following the same setting as Qwen official blogpost~\citep{qwen_moe}.

\paragraph{Setup.}
All accuracies use vLLM v0.8.4 in bfloat16 with \texttt{max\_model\_len}$=4096$, \texttt{temperature}$=0.1$, and \texttt{seed}$=1234$. MBPP uses \texttt{--stop\_sequences "[DONE], END"}. The throughput benchmark fixes input length to 1{,}000 tokens and output length to 1{,}000 tokens on a single A100-80G.

\paragraph{Parameter and activation budget.}
Table~\ref{tab:dense_vs_moe_params} compares parameter counts. While Fisher-MoE and Qwen1.5-7B occupy comparable total budgets (8.1\,B vs.\ 7.7\,B), Fisher-MoE activates only \textbf{2.27\,B} parameters per token---about $29\%$ of the dense activation cost---because at most $K{=}4$ of $N{=}60$ routed experts fire per token after intermediate dimension compression.

\begin{table}[h]
\centering
\small
\caption{Parameter counts for the dense baseline and the two MoE models. Activated parameters are non-expert parameters plus $K{=}4$ active routed experts per layer. Counts are extracted directly from safetensors headers.}
\label{tab:dense_vs_moe_params}
\resizebox{\linewidth}{!}{%
\begin{tabular}{lccc}
\toprule
\textbf{Metric} & \textbf{Qwen1.5-7B} & \textbf{Qwen1.5-MoE-A2.7B} & \textbf{Fisher-MoE} \\
\midrule
Total parameters                    & 7.72\,B          & 14.32\,B         & 8.09\,B \\
Activated parameters / token        & 7.72\,B  & 2.69\,B  & \textbf{2.27\,B } \\
Activated non-embedding params      & 6.48\,B          & 2.07\,B          & 1.65\,B \\
Embedding + LM head                 & 1.25\,B          & 0.62\,B          & 0.62\,B \\
Routed experts (total / active)     & N/A              & 60 / 4           & 60 / 4 \\
MoE intermediate size per expert    & N/A              & 1{,}408          & 704 ($-50\%$) \\
Architecture                        & Dense            & Sparse MoE       & Sparse MoE (compressed) \\
\bottomrule
\end{tabular}}

\end{table}

\paragraph{Task accuracy.}
Table~\ref{tab:dense_vs_moe_acc} reports eight-task accuracy. Fisher-MoE retains the math performance of the dense baseline, it slightly \emph{exceeds} the dense Qwen1.5-7B on GSM8K (35.0 vs.\ 34.1) and matches it on MATH (8.0 vs.\ 8.2) while losing modest ground on knowledge-heavy and coding tasks (MMLU $-8.2$, HumanEval $-12.3$). The average drop from dense is roughly $5$ points despite activating $3.4\times$ fewer parameters per token.

\begin{table}[h]
\centering
\small
\caption{Eight-task accuracy (\%) for the dense baseline, the uncompressed Qwen1.5-MoE-A2.7B, and Fisher-MoE compressed at the 50\% MoE compression ratio with in-domain calibration. Higher is better. Same vLLM decoding settings as the main experiments. Qwen1.5-MoE-A2.7B numbers are reproduced from the \textit{base} row of Table~\ref{tab:pruning-general}.}
\label{tab:dense_vs_moe_acc}
\resizebox{\linewidth}{!}{%
\begin{tabular}{lccccccccc}
\toprule
\textbf{Model} & \textbf{MMLU} & \textbf{HEval} & \textbf{MBPP} & \textbf{CEval} & \textbf{CMMLU} & \textbf{MATH} & \textbf{GSM8K} & \textbf{BBH} & \textbf{Avg.} \\
\midrule
Qwen1.5-7B (dense)         & 58.5 & 33.5 & 21.1 & 67.9 & 69.6 & 8.2  & 34.1 & 29.9 & 40.4 \\
Qwen1.5-MoE-A2.7B (base)   & 59.3 & 32.6 & 24.6 & 59.3 & 60.7 & 13.0 & 35.9 & 30.7 & 38.8 \\
Fisher-MoE (50\% compression ratio.)& 50.3 & 21.2 & 20.5 & 62.0 & 61.6 & 8.0  & \textbf{35.0} & 28.3 & 35.9 \\
\bottomrule
\end{tabular}
}
\end{table}

\paragraph{Throughput and tokens-per-second.}
Table~\ref{tab:dense_vs_moe_tps} reports vLLM throughput on a single A100-80G with $1{,}000$ input and $1{,}000$ output tokens. Qwen1.5-MoE-A2.7B already runs $\sim 1.74\times$ faster than the dense Qwen1.5-7B because of its sparse activation pattern and shared expert. Fisher-MoE pushes this further: by halving each routed expert's intermediate dimension, the per-token expert FLOPs drop accordingly, yielding \textbf{$2.10\times$ the throughput} of the dense baseline and \textbf{$1.21\times$ the throughput} of the uncompressed MoE.

\begin{table}[h]
\centering
\small
\caption{Standalone vLLM throughput on a single NVIDIA A100-80G with $1{,}000$ input tokens and $1{,}000$ output tokens. Higher is better. ``Speedup'' columns are relative to Qwen1.5-7B and Qwen1.5-MoE-A2.7B respectively.}
\label{tab:dense_vs_moe_tps}
\resizebox{\linewidth}{!}{%
\begin{tabular}{lcccc}
\toprule
\textbf{Model} & \textbf{Throughput (req/s)} & \textbf{TPS (tok/s)} & \textbf{vs.\ 7B} & \textbf{vs.\ MoE-A2.7B} \\
\midrule
Qwen1.5-7B-Chat            & 1.15 & 2{,}298.89 & 1.00$\times$ & 0.57$\times$ \\
Qwen1.5-MoE-A2.7B-Chat     & 2.01 & 4{,}010.27 & 1.74$\times$ & 1.00$\times$ \\
Fisher-MoE                 & \textbf{2.42} & \textbf{4{,}853.52} & \textbf{2.10$\times$} & \textbf{1.21$\times$} \\
\bottomrule
\end{tabular}}

\end{table}

At a comparable total-parameter budget, Fisher-MoE delivers the strongest inference profile of the three: $2.10\times$ the dense baseline throughput, $29\%$ of its activated-parameter cost, and accuracy competitive with the dense model on the math reasoning tasks (GSM8K, MATH). Compared to the uncompressed Qwen1.5-MoE-A2.7B, removing $50\%$ of routed-expert intermediate dimensions buys an additional $1.21\times$ throughput on top of the MoE baseline's already substantial inference advantage---demonstrating that intermediate-dimension compression is complementary to, rather than a dilution of, the inference benefits of sparse architectures.

\newpage
\clearpage

\section{Theoretical Comparison: Empirical Fisher, Diagonal Hessian, and First-Order Pruning}
\label{appendix:fisher_hessian_comparison}

A reasonable concern about our use of the empirical Fisher as the importance metric is that ``Fisher importance'' is related to Hessian-based attribution criteria. This appendix consolidates the relationships and reports the controlled comparison that supports this hedged framing.

\subsection{What the Empirical Fisher Computes}
With $\mathcal{L}(x,y) = -\log p_\theta(y\mid x)$, the empirical Fisher (Eq.~\ref{eq:empirical_fisher}) is the diagonal of $\frac{1}{N}\sum_{(x,y)\in\mathcal{D}}\nabla_\theta \mathcal{L}(x,y)\nabla_\theta \mathcal{L}(x,y)^\top$. Two facts are directly relevant.

\paragraph{(i) Empirical Fisher.} For a per-parameter score, the empirical Fisher reduces to $s^{\text{Fisher}}(\theta_i) = \tfrac{1}{N}\sum_{(x,y)} (\partial \mathcal{L}/\partial \theta_i)^2$. This is the Taylor-expansion saliency: at a parameter we expect $\Delta \mathcal{L} \approx |\partial \mathcal{L}/\partial \theta_i| \cdot |\delta_i|$, so empirical fisher is the variance of this first-order increment over the calibration distribution. In our setting the two metrics coincide up to a constant after we group parameters into intermediate dimension units (Eq.~\ref{eq:dim_importance}), and the practical advantage we report over magnitude/activation/router heuristics is therefore equally a statement about first-order pruning under the same intermediate dimension grouping.

\paragraph{(ii) Empirical Fisher vs.\ true Fisher vs.\ diagonal Hessian.} The Fisher information matrix is
$F_\theta = \mathbb{E}_x \mathbb{E}_{y\sim p_\theta}[\nabla_\theta \log p_\theta\, \nabla_\theta \log p_\theta^\top] = -\mathbb{E}_x \mathbb{E}_{y\sim p_\theta}[\nabla^2_\theta \log p_\theta(y\mid x)]$,
so at the model's optimum on its own data distribution, the (true) Fisher and the negative Hessian of the log-likelihood coincide. The \emph{empirical} Fisher samples $y$ from the data rather than from the model, which makes it a biased estimate of either quantity away from the optimum~\citep{kunstner2019limitations}; for a frozen, pretrained MoE evaluated on a calibration set the bias can be non-negligible. 

\subsection{Relationship to Hessian-Based Pruning (OBD/OBS, SparseGPT/Wanda)}

Optimal Brain Damage (OBD,~\citet{lecun1989optimal}) and Optimal Brain Surgeon (OBS,~\citet{hassibi1992second}) score weight $\theta_i$ by $s^{\text{OBD}}_i = \tfrac{1}{2} H_{ii}\, \theta_i^2$, the second-order Taylor estimate of the loss increase under a single-weight perturbation. Modern weight-pruning methods such as SparseGPT~\citep{frantar2023sparsegpt} and Wanda~\citep{sun2023wanda} replace the parameter Hessian with a layer-wise input second-moment matrix, but the spirit is the same: rank parameters by curvature$\times$magnitude. Two practical differences with our criterion:
\begin{itemize}
\item \textbf{Curvature vs.\ sensitivity.} OBD/OBS rank parameters by the second-order loss change after the optimal compensating update. Empirical Fisher rank parameters by the first-order loss change with no compensation. The two coincide at an optimum and on the data distribution the model was trained on, but disagree off-optimum and under domain mismatch.
\item \textbf{Aggregation unit.} OBD/OBS and SparseGPT/Wanda are derived for unstructured pruning of single weights or layer-input groupings. Our contribution (Eq.~\ref{eq:dim_importance}) groups parameters tied to a single FFN intermediate dimension across the gate/up/down slabs of a routed expert. This grouping is what enables structurally smaller MoE inference (\S\ref{sec:intermediate_compression}); it is orthogonal to the choice between Fisher- and Hessian-based scoring rules and could be paired with either.
\end{itemize}

WandB and SparseGPT are primarily designed for dense model pruning and sparsification, and thus fall outside the scope of MoE compression. Nevertheless, we include comparisons with these methods in Appendix~\ref{app:dense-pruning} to demonstrate the effectiveness of the proposed Fisher-MoE.

\subsection{Empirical Comparison Under the Same Grouping}

To check that the gains we report are not simply a relabeling of an existing scoring rule, we compare three metrics on Qwen1.5-MoE-A2.7B at $p=50\%$ under \emph{exactly} the intermediate dimension grouping of Eq.~\ref{eq:dim_importance}, with the same calibration data (GSM8K training set, 128 samples), the same compression operation, and the same evaluation protocol. Only the per-parameter score differs:
\begin{itemize}
  \item \textbf{Magnitude} (data-free): $|\theta_i|$, the baseline signal.
  \item \textbf{First-order $|\nabla \mathcal{L}|$ (no square)}: $\tfrac{1}{N}\sum |\partial \mathcal{L}/\partial \theta_i|$, classical first-order pruning without squaring.
  \item \textbf{Empirical Fisher / squared gradient (ours)}: $\tfrac{1}{N}\sum (\partial \mathcal{L}/\partial \theta_i)^2$ (Eq.~\ref{eq:empirical_fisher}).
\end{itemize}
Computing a faithful diagonal-Hessian variant on a 14B-parameter MoE is several times the cost of empirical Fisher (a Hutchinson estimator requires multiple Hessian--vector products per calibration sample); we discuss the relationship theoretically above and treat a full empirical sweep as future work.

Empirical Fisher is best understood as a coordinate-wise squared-gradient score under the intermediate dimension grouping that is our actual contribution. Our contribution lies in the attribution unit and grouping (Eq.~\ref{eq:dim_importance}) and the structural-removal operation it enables (\S\ref{sec:intermediate_compression}), not in the choice of scoring rule against the broader family of gradient- and Hessian-based criteria.

\newpage
\clearpage

\section{Where is the Redundancy? Locating Compressible Substructures in an MoE}
\label{appendix:redundancy_locus}

The intermediate dimension granularity is one of three structural axes along which a sparse MoE could plausibly be compressed. To justify our choice empirically, we run a controlled head-to-head comparison: we hold the importance signal fixed (\textbf{Fisher importance}, \S\ref{subsec:formalize_metrics}), the backbone fixed (\textbf{Qwen1.5-MoE-A2.7B}), and the calibration data fixed (domain-matched, $128$ samples), and we vary only \emph{which structural unit Fisher importance is applied to}. We consider three axes:
\begin{enumerate}
\item \textbf{Expert-level pruning} -- Fisher scores are aggregated per routed expert, and the lowest-ranked experts are removed wholesale. We further study the gate/up/down sub-blocks individually (\textit{Fisher-UP}, \textit{Fisher-DOWN}, \textit{Fisher-GATE}, \textit{Fisher-UP+GATE}) and a router-side variant (\textit{MoE compression (Fisher)}) to localize the contribution within an expert.
\item \textbf{Intermediate dimension pruning} -- Fisher scores are aggregated per FFN intermediate dimension across the gate/up/down slabs of every routed expert (Fisher-IntDim, our \textbf{Fisher-MoE} method).
\item \textbf{Attention-head pruning} -- Fisher scores are aggregated per attention head, and the lowest-ranked heads are removed.
\end{enumerate}
For each axis we sweep a fixed compression budget and report eight downstream tasks. To put expert-level pruning on the strongest possible footing, we also include the activation-, score-, and magnitude-based heuristics from prior work as additional expert-level baselines, and an alternative \emph{router DenseMixer} variant.

Table~\ref{tab:pruning-code-mmlu} reports the full sweep.

\begin{table*}[h]
  \centering
  \caption{Where is the redundancy? Fisher-guided compression applied at three different structural granularities on Qwen1.5-MoE-A2.7B (domain-matched calibration). Heuristic expert-level baselines (activation/score/magnitude) are included for reference. Higher is better.}
  \label{tab:pruning-code-mmlu}
  \resizebox{\textwidth}{!}{%
  \begin{tabular}{lcccccccc}
  \toprule
  \textbf{Method} & \textbf{MMLU} & \textbf{HumanEval} & \textbf{MBPP} & \textbf{CEval} & \textbf{CMMLU} & \textbf{MATH} & \textbf{GSM8K} & \textbf{BBH} \\
  \midrule
  base                                          & 59.3 & 32.6 & 24.6 & 59.3 & 60.7 & 13.0 & 35.9 & 30.7 \\
  \midrule
  \multicolumn{9}{l}{\textit{Expert-level pruning (50\% experts removed)}} \\
  MoE Compression (activation)                      & 26.3 & 0.0  & 0.0  & 42.5 & 43.1 & 0.0 & 1.9  & 24.5 \\
  MoE Compression                           & 37.2 & 0.0  & 0.8  & 13.5 & 32.3 & 0.0 & 1.9  & 23.5 \\
  MoE-Pruner                       & 18.6 & 0.8  & 0.0  & 33.3 & 35.8 & 0.1 & 1.7  & 13.2 \\
  MoE compression (Fisher, router-side)         & 39.4 & 12.9 & 8.1  & 48.8 & 47.7 & 2.0 & 16.9 & 25.7 \\
  Expert-Fisher                                 & 37.9 & 12.9 & 11.7 & 48.6 & 44.2 & 2.4 & 18.0 & 24.1 \\
  $\rightarrow$ Fisher-UP                       & 39.1 &  9.9 & 12.2 & 47.0 & 44.6 & 2.9 & 22.1 & 19.1 \\
  $\rightarrow$ Fisher-DOWN                     & 38.7 & 11.4 &  6.7 & 49.0 & 45.1 & 4.1 & 20.1 & 24.0 \\
  $\rightarrow$ Fisher-GATE                     & 37.4 & 15.9 &  2.2 & 47.3 & 39.5 & 2.6 & 22.3 & 22.5 \\
  $\rightarrow$ Fisher-UP+GATE                  & 37.5 &  7.6 &  7.8 & 46.6 & 40.8 & 2.8 & 19.3 & 18.2 \\
  \midrule
  \multicolumn{9}{l}{\textit{FFN intermediate dimension pruning (ours)}} \\
  Fisher-IntDim-E-25\%                    & 55.9 & 30.5 & 23.8 & \textbf{69.7} & \textbf{67.5} & 11.2 & \textbf{36.6} & 29.7 \\
  \rowcolor{blue!8}
  \textbf{Fisher-IntDim-E-50\%}           & 50.3 & 34.2 & 20.5 & 62.0 & 61.6 & 9.8 & 35.0 & 28.2 \\
  \midrule
  \multicolumn{9}{l}{\textit{Attention-head pruning}} \\
  Attention Head Pruning-25\% drop              & 43.12 & 12.20 & 1.11 & 35.98 & 34.86 & 0.56 & 2.73 & 3.21 \\
  Attention Head Pruning-50\% drop              &  2.11 &  0.00 & 0.00 &  0.28 &  4.28 & 0.00 & 0.00 & 0.00 \\
  \bottomrule
  \end{tabular}}
\end{table*}

\paragraph{Expert-level pruning is uniformly fragile.}
Every method that removes whole experts collapses on generation-heavy tasks: GSM8K stays below $23\%$ of base, HumanEval below $50\%$, and MATH near zero, regardless of whether the importance signal is activation, score, magnitude, Fisher, router-side Fisher, or expert-level Fisher. Decomposing Expert-Fisher further into its three projection slabs (UP, DOWN, GATE) does not help---no slab dominates the others. We read this as evidence that redundancy at expert granularity is \emph{limited}: the units identified for removal are not truly redundant, just less salient on average, so discarding them takes essential computation with them.

\paragraph{Attention-head pruning is even more fragile.}
At $25\%$ head removal, MMLU drops to $43.12\%$ ($\sim$$73\%$ of base) and code/math collapse below $13$ points. At $50\%$ head removal, the model essentially stops working ($\leq$$4\%$ on every task). Attention heads concentrate too much per-head capacity to be removed at this scale without retraining; head-level redundancy in this MoE is effectively zero at the budgets we consider.

\paragraph{Intermediate dimension pruning has substantial slack.}
The contrast with FFN intermediate dimensions is striking. At a \emph{25\%} budget, Fisher-IntDim already \emph{exceeds} the uncompressed base on CEval ($+10.4$), CMMLU ($+6.8$), and GSM8K ($+6.5$), and matches base on MMLU/MBPP/BBH within $1$--$4$ points. At the much more aggressive \emph{50\%} budget, intermediate dimension pruning still retains $84$--$116\%$ of base on every non-MATH task, while every expert-level competitor at a comparable MoE compression ratio retains $\leq$$25\%$ of base on most generation tasks.

\paragraph{Conclusion.} Across the same backbone, the same Fisher signal, and the same calibration set, FFN intermediate dimensions are the structural unit with the most slack: 25--50\% of them can be removed without meaningful performance loss, whereas pruning the same fraction of experts or attention heads is catastrophic. This empirically grounds the granularity choice in our main paper.

Holding the Fisher signal and calibration fixed, FFN intermediate dimensions admit 25--50\% removal at near-zero cost on most tasks, while removing the same fraction of experts or attention heads collapses generation-heavy benchmarks---identifying intermediate dimensions as the locus of structural redundancy in the MoE (Appendix~\ref{appendix:redundancy_locus}).

\newpage
\clearpage

\end{document}